\documentclass[11pt]{article}

\usepackage[final]{acl}

\usepackage{times}
\usepackage{latexsym}

\usepackage[utf8]{inputenc}

\usepackage{microtype}

\usepackage{inconsolata}

\usepackage{graphicx}

\usepackage{times}  
\usepackage{helvet}  
\usepackage{courier}  
\usepackage{graphicx} 
\usepackage{natbib}  
\usepackage{caption} 
\usepackage{algorithm}
\usepackage{algorithmic}
\usepackage{amsmath}
\usepackage{amssymb}
\usepackage{tikz}
\usetikzlibrary{positioning,trees,fit}
\usepackage{colortbl}
\usepackage{subcaption}
\usepackage{xcolor}
\usepackage{natbib}
\usepackage{graphicx}
\usepackage{soul}     
\usepackage{colortbl}
\usepackage{pgfplots}
\pgfplotsset{compat=1.18}
\usepackage{array}

\sethlcolor{gray!30}

\usepackage{newfloat}
\usepackage{listings}
\usepackage{booktabs}
\usepackage{geometry}

\title{ClimateCause: Complex and Implicit Causal Structures in Climate Reports}

\author{
    \textbf{Liesbeth Allein\textsuperscript{\rm $\diamondsuit$}\textsuperscript{\rm $\dagger$}}\thanks{Corresponding author: \url{Liesbeth.Allein@UGent.be}.},
    \textbf{Nataly Pineda-Castañeda\textsuperscript{\rm $\ddagger$}},
    \textbf{Andrea Rocci\textsuperscript{\rm $\ddagger$}},
    \textbf{Marie-Francine Moens\textsuperscript{\rm $\diamondsuit$}}
    \\
    \\
    \textsuperscript{$\diamondsuit$}Department of Computer Science, KU Leuven, Belgium \\
    \textsuperscript{$\dagger$}Department of Electronics and Information Systems, Ghent University, Belgium \\
      \textsuperscript{$\ddagger$}Institute of Argumentation, Linguistics, and Semiotics (IALS), \\ Università della Svizzera italiana, Switzerland
}

\begin{document}

\maketitle

\begin{abstract}

Understanding climate change requires reasoning over complex causal networks. Yet, existing causal discovery datasets predominantly capture explicit, direct causal relations. We introduce ClimateCause, a manually expert-annotated dataset of higher-order causal structures from science-for-policy climate reports, including implicit and nested causality. Cause-effect expressions are normalized and disentangled into individual causal relations to facilitate graph construction, with unique annotations for cause-effect correlation, relation type, and spatiotemporal context. We further demonstrate ClimateCause's value for quantifying readability based on the semantic complexity of causal graphs underlying a statement. Finally, large language model benchmarking on correlation inference and causal chain reasoning highlights the latter as a key challenge.

\end{abstract}

\section{Introduction}

\begin{table*}[ht!]
    \centering
    \tiny
    \begin{tabular}{p{3.5cm}|p{1.6cm}>{\columncolor{gray!15}}p{1.7cm}|p{0.05cm}p{0.3cm}p{0.05cm}p{0.05cm}p{0.05cm}>{\columncolor{gray!15}}p{0.05cm}p{0.05cm}|p{0.1cm}>{\columncolor{gray!15}}p{0.1cm}p{0.1cm}|p{0.1cm}>{\columncolor{gray!15}}p{0.1cm}|p{0.1cm}p{0.1cm}>{\columncolor{gray!15}}p{0.1cm}}
    \cmidrule{2-18}
    \multicolumn{1}{c}{} & \multicolumn{2}{|c}{\textbf{Scope \& Coverage}} & \multicolumn{7}{|c}{\textbf{Causal Structures}} & \multicolumn{3}{|c}{\textbf{Properties}} & \multicolumn{2}{|c}{\textbf{Context}} & \multicolumn{3}{|c|}{\textbf{Reliability}} \\
    \cmidrule{2-18}
        \textbf{Dataset} & \textbf{Topic} & \textbf{Source} & \rotatebox{90}{\textbf{Presence \& Abs.}} & \rotatebox{90}{\textbf{\# CR}} & 
\rotatebox{90}{\textbf{Cross-Sent. CR}} & 
\rotatebox{90}{\textbf{Trigger}} & 
\rotatebox{90}{\textbf{Consist. Formul.}} & 
\rotatebox{90}{\textbf{Nested CS}} & 
\rotatebox{90}{\textbf{Complex CS}} & 
\rotatebox{90}{\textbf{Implicit CR}} & 
\rotatebox{90}{\textbf{Correlation}} & 
\rotatebox{90}{\textbf{Relation type}} & 
\rotatebox{90}{\textbf{In Discourse}} & 
\rotatebox{90}{\textbf{In Time/Space}} & 
\rotatebox{90}{\textbf{Manual Annot.}} & 
\rotatebox{90}{\textbf{Expert Annot.}} & 
\rotatebox{90}{\textbf{Scientific CRs}} \\
    \midrule
    BioCause \citep{mihuailua2013biocause} & Infectious diseases & Biomedical articles & \checkmark & 851 & \checkmark & \checkmark & & &   & \checkmark &  &  & \checkmark &  & \checkmark & \checkmark & \checkmark \\
    Causal-TimeBank \citep{mirza-etal-2014-annotating} & Varia & News articles & \checkmark & 318 &  & \checkmark & & &  &  &  & \checkmark &  &  & \checkmark & \checkmark &  \\
    CaTeRS \citep{mostafazadeh-etal-2016-caters}  & Common sense & Everyday stories & \checkmark & 308 & \checkmark &  & & &   & \checkmark &  & \checkmark & \checkmark &  & \checkmark & \checkmark &  \\
    BECauSE 2.0 \citep{dunietz-etal-2017-corpus} & Varia & News articles, Penn Treebank, Congress hearings & \checkmark & 1,803 &  & \checkmark & & &   & \checkmark &  & \checkmark &  &  & \checkmark & \checkmark &  \\
    CRAB \citep{romanou2023crab} & Varia & News articles &  & 2,730 & \checkmark &  & \checkmark & &  \checkmark &  &  &  & \checkmark &  & \checkmark &  &  \\
    CNC \citep{tan2022causal} & Socio-political events & News articles & \checkmark & 1,957 &  & \checkmark & & &   &  &  &  & \checkmark &  & \checkmark & \checkmark &  \\
   RECESS \citep{tan-etal-2023-recess} & Socio-political events & News articles & \checkmark & 2,754 &  & \checkmark & & &   &  &  &  & \checkmark & & \checkmark & \checkmark &  \\
   CCNC \citep{hagen2025investigating} & Varia & News articles & \checkmark & 3,415 & & \checkmark & & & & \checkmark & & & \checkmark & & \checkmark & &  \\
   ACCESS \citep{vo-etal-2025-access} & Common sense & Stories &  & 1,494 & \checkmark &  & \checkmark & & \checkmark  &  &  &  & \checkmark &  & \checkmark &  &  \\
    PolarIs3CAUS \citep{pineda2025polaris3} & Climate change & Reddit &  & 95 &  & \checkmark & \checkmark & &  &  &  & \checkmark & \checkmark &  & \checkmark & \checkmark &  \\
    PolarIs4CAUS \citep{pineda2025polaris4} & Climate change & Twitter (X) &  & 181 &  & \checkmark & \checkmark & &  &  &  & \checkmark &  &  & \checkmark & \checkmark &  \\
    \midrule
    \textbf{\textit{ClimateCause} (ours)} & \textbf{Climate change} & \textbf{Science-for-policy reports} & \textbf{\checkmark} & \textbf{874} & \textbf{\checkmark} & \textbf{\checkmark} & \textbf{\checkmark} & \textbf{\checkmark} & \textbf{\checkmark} & \textbf{\checkmark} & \textbf{\checkmark} & \textbf{\checkmark} & \textbf{\checkmark} & \textbf{\checkmark} & \textbf{\checkmark} & \textbf{\checkmark} & \textbf{\checkmark} \\
    \bottomrule
    \end{tabular}
    \caption{Comparison to existing resources, highlighting (nearly) unique features of \textit{ClimateCause} in gray.}
    \label{tab:comparison-against-existing-datasets}
\end{table*}

Causality is a fundamental driver of climate change and climate change discourse. Climatic phenomena unfold within complex causal networks, where causal relations are further complicated by contextual factors that introduce uncertainty, signal confounders, and highlight the variability of causal strength and direction \citep{pearl2009causality,yarlett2019uncertainty,cui-etal-2025-uncertainty}. For example, \textit{why would an increase in global temperature of 1.5°C lead to more frequent floods in Africa but not in South America?} Causal reasoning also underlies policy making, where mitigation strategies related to climate change are designed through responsibility attribution and counterfactual analysis \citep{jang2013framing,jamieson2015responsibility,kalch2021responsible}. 

\begin{figure}[ht!]
    \centering
    \begin{minipage}{\linewidth}
    \scriptsize
    \textbf{Statement:} \textit{``Climate change has reduced food security and affected water security due to warming, changing precipitation patterns, reduction and loss of cryospheric elements, and greater frequency and intensity of climatic extremes, thereby hindering efforts to meet Sustainable Development Goals.''} \\
    \end{minipage}
    \begin{tikzpicture}[
      grow=right,
      sibling distance=1.05cm,
      level distance=1.15cm,
      every node/.style={
      anchor=west,
      text width=2.2cm,
      font=\scriptsize,
      draw,
      rectangle,
      rounded corners,
      fill=gray!15,
      align=center,
      inner sep=3pt},
      parent/.style={text width=1.1cm, font=\scriptsize},
      edge from parent path={
        [->] (\tikzparentnode.east) -- ++(0,0) -- (\tikzchildnode.west)
      }
    ]

    \node[parent] (root) {climate change}
      child {node (n1) {greater intensity of climatic extremes}}
      child {node (n2) {greater frequency of climatic extremes}}
      child {node (n3) {loss of cryospheric elements}}
      child {node (n4) {reduction of cryospheric elements}}
      child {node (n5) {changing precipitation patterns}}
      child {node (n6) {warming}};

    \draw[->, very thick] (root.east) -- (n6.west)
      node[midway, above, xshift=-0.8cm, yshift=0.2cm,
           font=\scriptsize, text width=1.5cm,
           draw, dashed, align=center, inner sep=2pt, line width=0.4pt, rounded corners, fill=none] 
      {\textbf{\textit{correlation}}: + \\ \textbf{\textit{relation type}}: + \\ \textbf{\textit{explicit}}: E};

    \coordinate (sharedBase) at ([xshift=0.7cm]n4.east);
    \node[anchor=west,text width=0.8cm,font=\scriptsize] at (sharedBase) (c1) {reduced food security};
    \node[anchor=west,text width=0.8cm,font=\scriptsize] at ([yshift=-0.8cm]c1.south west) (c2) {affected water security};

    \foreach \n in {n1,n2,n3,n4,n5,n6} {
      \draw[->] (\n.east) -- (c1.west);
      \draw[->] (\n.east) -- (c2.west);
    }

    \node[anchor=west,text width=1.1cm,font=\scriptsize] at ([xshift=0.5cm,yshift=-0.65cm]c1.east) (c3) {hindered efforts to meet SDGs};

    \draw[->] (c1.east) -- (c3.west);
    \draw[->] (c2.east) -- (c3.west);

    \draw[->, very thick] (n1.east) -- (c2.west)
      node[midway, above, font=\scriptsize, text width=2cm, xshift=1.2cm, yshift=-0.9cm, draw, dashed, align=center, inner sep=2pt, line width=0.4pt, rounded corners, fill=none] {\textbf{\textit{correlation}}: -- \\ \textbf{\textit{relation type}}: + \\ \textbf{\textit{explicit}}: E};

    \end{tikzpicture}
    \caption{A sample from the \textit{ClimateCause} dataset, showcasing the complex causal graphs and fine-grained annotations it contains.}
    \label{fig:complex_causal_graph_introduction}
\end{figure}
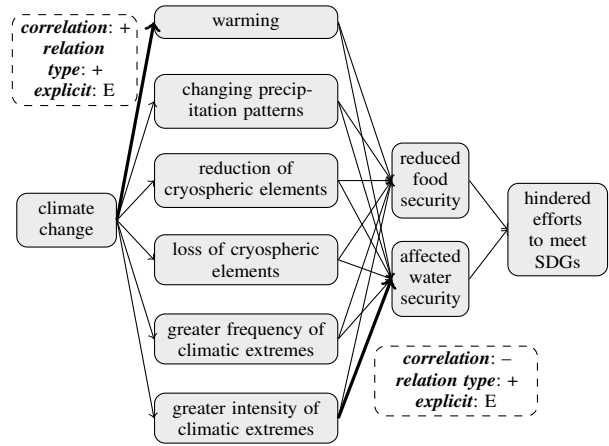

Yet, existing datasets for causal discovery from text lack the granularity and abstraction for domains characterized by such complex causality \citep{mihuailua2013biocause,mirza-etal-2014-annotating,mostafazadeh-etal-2016-caters,dunietz-etal-2017-corpus,romanou2023crab,tan2022causal,vo-etal-2025-access,pineda2025polaris3,pineda2025polaris4}. They primarily capture explicit direct cause-effect relations and omit those that are implicitly reported through word and sentence semantics; e.g., \textit{``\underline{anthropogenic} greenhouse gas emissions''} evokes causal relation \textit{humans} $\rightarrow$ \textit{greenhouse gas emissions}. Since most are sourced from news and social media, they typically report singular rather than general causality (e.g., the effects of a specific hurricane) and do not discuss higher-order, complex causal structures as commonly done in science-related discourse. 

This paper introduces\textbf{ \textit{ClimateCause}, a manually-annotated dataset for causal discovery of complex and implicit causal structures from climate change reports} (Figure \ref{fig:complex_causal_graph_introduction})\footnote{Data and code are publicly available: \url{https://github.com/laallein/ClimateCause}.}. Unlike prior resources, cause and effect representations are manually normalized through noun phrase reformulation, and multi-event representations are disentangled to facilitate graph construction. Implicit and nested causal expressions are captured and each causal relation is further annotated with cause-effect correlation, relation type, and spatiotemporal context to create semantically-rich causal graphs.

Analyses on \textit{ClimateCause} provide insights into the dominant semantic properties of the causation framing used in climate change reports. These reports are generally marked by low readability \citep{barkemeyer2016linguistic} and complex terminology \citep{bruine2021public}. We take a causality-centered perspective on readability and propose several metrics to quantify the complexity of reported causality based on semantic properties of underlying causal graphs. We further showcase the potential of \textit{ClimateCause} for benchmarking the causal reasoning abilities of LLMs through correlation inference and causal chain reasoning.

\section{Comparison to Existing Resources} \label{sec:related_work}

We compare \textit{ClimateCause} against existing text-based causal discovery datasets, in which cause and effect are expressed as textual spans. This ensures a fair comparison among datasets explicitly designed for extracting causal relations directly from text only. In that respect, we exclude datasets with multimodal data \citep{liang2025videvent,shen-etal-2025-exploring} or causal questions in Q\&A \citep{bondarenko-etal-2022-causalqa,ceraolo2024analyzing,chi2024unveiling} and NLI setups \citep{roemmele2011choice,du-etal-2022-e,miliani2025explica}, and datasets that are designed to trigger causal reasoning strategies for answer selection \citep{jin2023cladder,jincan,sheth-etal-2025-causalgraph2llm}. We also omit causal knowledge graphs \citep{heindorf2020causenet,10.5555/3491440.3491942,9706608} and knowledge graphs that include causal relations as one of the relations \citep{speer2017conceptnet,sap2019atomic,hwang2021comet}. 

\textit{ClimateCause} is unique in its annotations for correlation, spatiotemporal context, and nested causality, establishing first-of-its-kind resources for investigating these three aspects for causal discovery (Table~\ref{tab:comparison-against-existing-datasets}). Alongside BioCause \citep{mihuailua2013biocause}, it is one of the few built from scientific texts.
Moreover, \textit{ClimateCause} includes a larger number of causal relations than related climate change-focused datasets \citep{pineda2025polaris3,pineda2025polaris4}.

\section{Dataset Construction: Source}

\paragraph{Data source and extraction} All texts in \textit{ClimateCause} are sourced from science-for-policy reports published by the Intergovernmental Panel on Climate Change (IPCC)\footnote{\url{https://www.ipcc.ch}}. These authoritative reports synthesize the scientific consensus on causes, impacts, risks, and mitigation of climate change. They serve a dual purpose: \textit{inform} non-expert stakeholders (i.e., policy-makers) about the scientific knowledge gained on a certain topic with confidence levels reflecting the level of consensus, while \textit{influencing} them. A citizen science initiative led by Semantic Climate\footnote{\url{https://semanticclimate.github.io/p/en/}} is transforming the IPCC reports into a structured knowledge graph. At retrieval time (May 2025), the AR6 Synthesis Report: Climate Change 2023\footnote{\url{https://www.ipcc.ch/report/ar6/syr/}} had been partially converted into a machine-readable format. We extracted 75 statements via SPARQL from a dedicated Wikibase instance\footnote{\url{https://kg-ipclimatec-reports.wikibase.cloud/wiki/Main_Page}}; \textbf{[\textit{Statement}] (\textit{string})}. These statements were selected based on the availability of confidence labels such that each statement reflects causal content that is grounded in a documented level of scientific consensus.
We rely on the statement boundaries defined by Semantic Climate, resulting in statements of one or more sentences. 
A breakdown of the report sections from which the statements are drawn is in Appendix \ref{appendix:ipcc_sections}.

\paragraph{Metadata} For each statement, we retrieve the Wikibase link to the statement \textbf{[\textit{Statement link}] (\textit{URL})}, section title \textbf{[\textit{Section}] (\textit{string})}, paragraph identifier \textbf{[\textit{Paragraph}] (\textit{id})}, ordinal position of the statement within the paragraph \textbf{[\textit{Series ordinal}] (\textit{integer})}, and associated confidence level \textbf{[\textit{Confidence level}] (\textit{label})}. 

\section{Dataset Construction: Annotation}  

\begin{table}[t]
    \centering
    \scriptsize
    \setlength\extrarowheight{-14pt}
    \begin{tabular}{p{0.5cm}p{1.6cm}p{4.4cm}}
    \toprule
    \multicolumn{2}{l}{\textbf{Feature}} & \textbf{Value} \\
    \midrule
    \multicolumn{2}{l}{\textit{Statement link}} & https://kg-ipclimatec-reports.wikibase.cloud/entity/statement/ Q31-91700757-4207-5de4-c0c1-5682b1be9db0 \\
    \multicolumn{2}{l}{\textit{Section}} & 2.1.2 Observed Climate System Changes and Impacts to Date \\
    \multicolumn{2}{l}{\textit{Paragraph}} & 2.1.2.c \\
    \multicolumn{2}{l}{\textit{Series ordinal}} & 1 \\
    \multicolumn{2}{l}{\textit{Statement}} & \textit{Climate change has caused substantial damages, and increasingly irreversible losses, in terrestrial, freshwater, cryospheric and coastal and open ocean ecosystems.} \\
    \multicolumn{2}{l}{\textit{Causation}} & Yes \\
    \multicolumn{2}{l}{\textit{Target}} & \textit{has caused} \\
    \multicolumn{2}{l}{\textit{Explicitness}} & Explicit \\
    \textit{Cause} & \textit{-- NP} & \textit{climate change} \\
     & \textit{-- Context} & / \\
     & \textit{-- No\_Quantifier} & climate change \\
     & \textit{-- Belongs\_to} & / \\
    \textit{Effect} & \textit{-- NP} & \textit{increasingly irreversible losses in freshwater ecosystems} \\
     & \textit{-- Context} & / \\
     & \textit{-- No\_Quantifier} & irreversible losses in freshwater ecosystems \\
     & \textit{-- Belongs\_to} & / \\
    \multicolumn{2}{l}{\textit{Relation type}} & Positive \\
    \multicolumn{2}{l}{\textit{Correlation}} & Positive \\
    \multicolumn{2}{l}{\textit{Abbreviations}} & / \\
    \multicolumn{2}{l}{\textit{Combined}} & / \\
    \multicolumn{2}{l}{\textit{Confidence level}} & High confidence \\
    \multicolumn{2}{l}{\textit{Nested}} & / \\
    \bottomrule
    \end{tabular}
    \caption{A sample from \textit{ClimateCause}.}
    \label{tab:data_sample}
\end{table}

\subsection{Annotation Features}
Table \ref{tab:data_sample} shows a sample from \textit{ClimateCause}, reflecting all annotation features of one cause-effect pair in a statement. We go over the features and illustrate them with samples from the dataset. Detailed annotation guidelines are included in Appendix \ref{appendix:annotation_guidelines}.

\subsubsection{Presence/Absence of Causality} \label{sec:presence_absence_causality}
\textbf{[\textit{Causation}] (\textit{yes/no})} marks whether a statement reports at least one causal relation. If not, no further action is taken for that statement.

\subsubsection{Target Word and Explicitness}\label{sec:target_word_implicit}
\textbf{[\textit{Target}] (\textit{string})} includes the target words in the statement that trigger the causal relation \citep{baker1998berkeley}. 
Examples of popular causation-evoking terms are \textit{``cause''} and \textit{``due to''}. We also include more subtle triggers to capture implicit causality. An example is \textit{``\underline{anthropogenic} greenhouse gas emissions''}, which evokes \textit{humans} $\rightarrow$ \textit{greenhouse gas emissions} through semantics. An explicit target may also be absent; [\textit{Target}] = \textit{/}. In that case, the reader needs to infer the causal relation. For example: \textit{warming} $\rightarrow$ \textit{rise in global mean sea level} in the statement \textit{``global mean sea level will rise by about 2–3 m if warming is limited to 1.5°C and 2–6 m if limited to 2°C.''}. The target guides the decision whether the causal relation is explicit (E) or implicit (I); \textbf{[\textit{Explicitness}] (\textit{E}/\textit{I})}.

\subsubsection{Cause and Effect} 
Annotators identify each cause-effect pair within a statement and resolve coreference where necessary. This way, we capture causal relations that would otherwise remain unidentified or cannot be understood outside their discourse context. We enforce consistency in cause and effect representation through (A) syntactic reformulation of events into noun phrases, (B) abbreviation resolution, and (C) disentanglement of multi-event events. We also identify (D) nested causal structures and (E) contextualize causal relations in time and space. 

\paragraph{(A) Syntactic reformulation} \textbf{[\textit{Cause--NP}] (\textit{string})} and \textbf{[\textit{Effect--NP}] (\textit{string})}, respectively, represent the noun phrase (NP) reformulation of cause and effect. We adopt the guidelines for NP reformulation from \citet{pineda2025polaris3,pineda2025polaris4}, which instructed annotators to minimally alter the original wording of cause and effect. For example: \textit{unsustainable agricultural expansion} $\rightarrow$ \textit{increased ecosystem vulnerability} in \textit{``Unsustainable agricultural expansion [...] increases ecosystem and human vulnerability''}. Including the target word \textit{``increases''} as an adjective in the effect is unlike dominant annotation approaches for semantic relations (e.g., \citet{baker1998berkeley}). Nonetheless, target inclusion arguably captures the effect more faithfully than exclusion, i.e., the cause leads to \textit{an increase in} and not \textit{the existence of ecosystem vulnerability}.

\paragraph{(B) Abbreviation resolution}
Annotators resolve the abbreviations used in cause and effect formulations directly in their NP reformulations and include them in \textbf{[\textit{Abbreviations}] (\textit{string})}. Abbreviations can be well-established, e.g., \textit{CO\textsubscript{2} = carbon dioxide}, or report-specific, e.g. \textit{AFOLU = \underline{A}griculture, \underline{F}orestry, and \underline{O}ther \underline{L}and \underline{U}se}. Appendix \ref{appendix:abbreviations} includes an overview of all abbreviations.

\paragraph{(C) Disentanglement of multi-event cause/effect} \label{sec:single_events}
Annotators decompose cause and effect into standalone events and actions. For example:
\\ \\
\begin{minipage}{\linewidth}
\scriptsize
    \textbf{Statement:} \textit{``Climate change has caused substantial damages, and increasingly irreversible losses, in terrestrial, freshwater, cryospheric and coastal and open ocean ecosystems.''}
\end{minipage}
\\
\begin{tikzpicture}[
  grow=right,
  sibling distance=0.4cm,
  level distance=1.1cm,
  every node/.style={anchor=west,text width=5.8cm,font=\scriptsize},
  parent/.style={text width=1cm, font=\scriptsize},
  edge from parent path={
    [->] (\tikzparentnode.east) -- ++(0,0) -- (\tikzchildnode.west)
  }
  ]

  \node[parent] {climate change}
    child {node {substantial damages in terrestrial ecosystems}}
    child {node {substantial damages in freshwater ecosystems}}
    child {node {substantial damages in cryospheric ecosystems}}
    child {node {substantial damages in coastal ecosystems}}
    child {node {substantial damages in open ocean ecosystems}}
    child {node {increasingly irreversible losses in terrestrial ecosystems}}
    child {node {increasingly irreversible losses in freshwater ecosystems}}
    child {node {increasingly irreversible losses in cryospheric ecosystems}}
    child {node {increasingly irreversible losses in coastal ecosystems}}
    child {node {increasingly irreversible losses in open ocean ecosystems}};

\end{tikzpicture}
\\
Disentangling causes and effects strongly contrasts with prior resources, which typically preserve mixed representations. Disentanglement has various advantages for causal graph building (e.g., event matching) and evaluation (e.g., one-by-one validation of cause-effect pairs). However, it may produce flawed or misleading causal graphs in case of (C.1) an elaboration of examples and (C.2) a combination of events.

\paragraph{(C.1) Elaboration of examples} A statement may elaborate on events that are examples of an overarching cause or effect. For example:
\\ \\
\begin{minipage}{\linewidth}
\scriptsize
    \textbf{Statement:} \textit{``Impacts in ecosystems from slow-onset processes such as ocean acidification, sea level rise or regional decreases in precipitation have also been attributed to human-caused climate change.''}
\end{minipage}
\\ \\
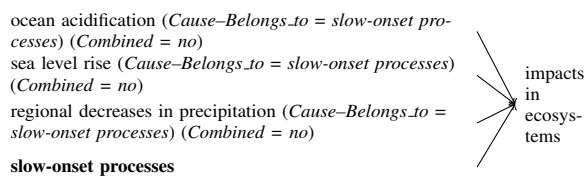
\begin{tikzpicture}[
  grow=right,
  sibling distance=0.5cm,
  level distance=0.3cm,
  every node/.style={anchor=west,text width=5cm,font=\scriptsize},
  parent/.style={text width=2cm, font=\scriptsize},
  edge from parent path={
    [->] (\tikzparentnode.east) -- ++(0,0) -- (\tikzchildnode.west)
  }
  ]

  \coordinate (sharedBase) at (0,0);
  \node[anchor=west,text width=6cm,font=\scriptsize] at (sharedBase) (c1) {ocean acidification (\textit{Cause--Belongs\_to} = \textit{slow-onset processes}) (\textit{Combined = no})};
  \node[anchor=west,text width=6cm,font=\scriptsize] at ([yshift=-0.2cm]c1.south west) (c2) {sea level rise (\textit{Cause--Belongs\_to} = \textit{slow-onset processes}) (\textit{Combined = no})};
  \node[anchor=west,text width=6cm,font=\scriptsize] at ([yshift=-0.25cm]c2.south west) (c3) {regional decreases in precipitation (\textit{Cause--Belongs\_to} = \textit{slow-onset processes}) (\textit{Combined = no})};
  \node[anchor=west,text width=6cm,font=\scriptsize] at ([yshift=-0.2cm]c3.south west) (c4) {\textbf{slow-onset processes}};

  \node[anchor=west,text width=0.8cm,font=\scriptsize] at ([xshift=0.5cm,yshift=0cm]c2.south east) (c5) {impacts in ecosystems };

  \draw[->] (c1.east) -- (c5.west);
  \draw[->] (c2.east) -- (c5.west);
  \draw[->] (c3.east) -- (c5.west);
  \draw[->] (c4.east) -- (c5.west);

\end{tikzpicture}
Representing the examples next to the overarching event/action (e.g., \textit{slow-onset processes}) in a causal graph without any indication of their subordinate relation to the overarching event may produce a misleading graph. The events may be wrongly seen as independent from each other. We mark such a subordinate relation between two events by including the overarching event in \textbf{[\textit{Cause--Belongs\_to}] (\textit{string})}, when the subordinate event is the cause, or in \textbf{[\textit{Effect--Belongs\_to}] (\textit{string})}, when the subordinate event is the effect. We also indicate that the causal relation does not 
arise from a combination of events  \textbf{[\textit{Combined}] (\textit{yes / \underline{no}})} to distinguish it from a combination of events.

\paragraph{(C.2) Combination of events} A causal relation may only arise from the combination of events. However, determining this faithfully would require extensive domain knowledge. Given the linguistic expertise of the annotators, we therefore focus on linguistic signals that indicate whether multiple causes produce a single effect or one cause leads to multiple effects. A clear example of such a signal is \textit{``the combined effects of''} in:
\\ \\
\begin{minipage}{\linewidth}
\scriptsize
    \textbf{Statement:} \textit{``Nearly 50\% of coastal wetlands have been lost over the last 100 years, as a result of the combined effects of localised human pressures, sea level rise, warming and extreme climate events.''}
\end{minipage}
\\

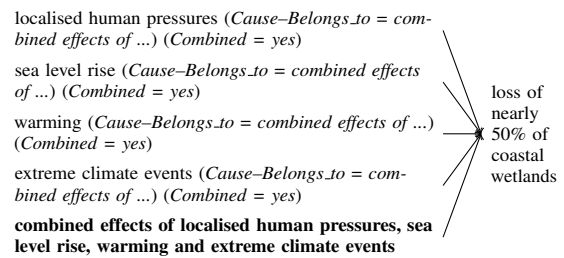
\begin{tikzpicture}[
  grow=right,
  sibling distance=0.5cm,
  level distance=0.2cm,
  every node/.style={anchor=west,text width=3.5cm,font=\scriptsize},
  parent/.style={text width=2cm, font=\scriptsize},
  edge from parent path={
    [->] (\tikzparentnode.east) -- ++(0,0) -- (\tikzchildnode.west)
  }
  ]

  \coordinate (sharedBase) at (0,0);
  \node[anchor=west,text width=5.5cm,font=\scriptsize] at (sharedBase) (c1) {localised human pressures (\textit{Cause--Belongs\_to} = \textit{combined effects of ...}) (\textit{Combined = yes})};
  \node[anchor=west,text width=5.5cm,font=\scriptsize] at ([yshift=-0.3cm]c1.south west) (c2) {sea level rise (\textit{Cause--Belongs\_to} = \textit{combined effects of ...}) (\textit{Combined = yes})};
  \node[anchor=west,text width=5.5cm,font=\scriptsize] at ([yshift=-0.3cm]c2.south west) (c3) {warming (\textit{Cause--Belongs\_to} = \textit{combined effects of ...}) (\textit{Combined = yes})};
  \node[anchor=west,text width=5.5cm,font=\scriptsize] at ([yshift=-0.3cm]c3.south west) (c4) {extreme climate events (\textit{Cause--Belongs\_to} = \textit{combined effects of ...}) (\textit{Combined = yes})};
  \node[anchor=west,text width=5.5cm,font=\scriptsize] at ([yshift=-0.3cm]c4.south west) (c6) {\textbf{combined effects of localised human pressures, sea level rise, warming and extreme climate events}};

  \node[anchor=west,text width=1.1cm,font=\scriptsize] at ([xshift=0.5cm,yshift=0cm]c3.east) (c5) {loss of nearly 50\% of coastal wetlands};

  \draw[->] (c1.east) -- (c5.west);
  \draw[->] (c2.east) -- (c5.west);
  \draw[->] (c3.east) -- (c5.west);
  \draw[->] (c4.east) -- (c5.west);
  \draw[->] (c6.east) -- (c5.west);

\end{tikzpicture}

We include the full noun phrase covering all events in the combination in [\textit{Cause--Belongs\_to}] (\textit{string}) or [\textit{Effect--Belongs\_to}] (\textit{string}) and mark that the causal relation is explicitly reported to arise from the combination of events [\textit{Combined}] (\textit{\underline{yes} / no}).

\paragraph{(D) Nested causality} \label{sec:embedded_causality}
Causal relations that are nested within another cause or effect are included as standalone causal relations. A striking example is the term \textit{CO2-FFI} (i.e., \textit{``carbon dioxide emissions from \underline{F}ossil \underline{F}uel combustion and \underline{I}ndustrial processes''}), which encapsulates \textit{fossil fuel combustion} $\rightarrow$ \textit{carbon dioxide emissions} and \textit{industrial processes} $\rightarrow$ \textit{carbon dioxide emissions}. 
We include the overarching effect 
in [\textit{Effect--Belongs\_to}] and mark the relation as \textbf{[\textit{Nested}] (\textit{\underline{yes}}/\textit{no})}.

\paragraph{(E) Spatiotemporal contextualization} \label{sec:contextualization_in_time_space}

The report often describes cause-effect pairs that occur in specific spatiotemporal contexts. We include such spatiotemporal information in \textbf{[\textit{Cause--Context}] (\textit{string})} and \textbf{[\textit{Effect--Context}] (\textit{string})}. The importance of contextualization is clear in the following statement, where different confidence levels have been assigned to the causality between events across geographical locations: 
\\ \\
\begin{minipage}{\linewidth}
\scriptsize
    \textbf{Statement:} \textit{``At 1.5°C global warming, heavy precipitation and flooding events are projected to intensify and become more frequent in most regions in Africa, Asia (high confidence), North America (medium to high confidence) and Europe (medium confidence).''}
\end{minipage}
\\ \\
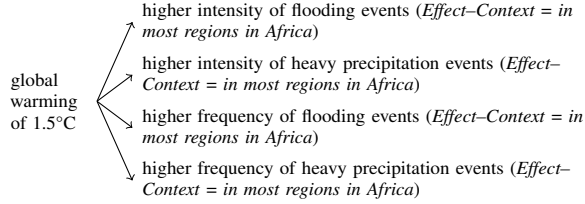
\begin{tikzpicture}[
  grow=right,
  sibling distance=0.7cm,
  level distance=1.1cm,
  every node/.style={anchor=west,text width=5.8cm,font=\scriptsize},
  parent/.style={text width=1cm, font=\scriptsize},
  edge from parent path={
    [->] (\tikzparentnode.east) -- ++(0,0) -- (\tikzchildnode.west)
  }
  ]

  \node[parent] {global warming of 1.5°C}
    child {node {higher frequency of heavy precipitation events (\textit{Effect--Context} = \textit{in most regions in Africa})}}
    child {node {higher frequency of flooding events (\textit{Effect--Context} = \textit{in most regions in Africa})}}
    child {node {higher intensity of heavy precipitation events (\textit{Effect--Context} = \textit{in most regions in Africa})}}
    child {node {higher intensity of flooding events (\textit{Effect--Context} = \textit{in most regions in Africa})}};

\end{tikzpicture}

These four causal relations are repeated four times, where the effect is linked to a different spatial context in [\textit{Effect--Context}], i.e., \textit{``in most regions in Asia''}, \textit{``in most regions in North America''}, and \textit{``in most regions in Europe''}. The relations are paired with their designated confidence level. We consider spatiotemporal information as contextual when it is not an essential aspect of the meaning of the cause or effect, which is in line with \textit{circumstances} in FrameNet \citep{baker1998berkeley}. We include it in [\textit{Cause--NP}] or [\textit{Effect--NP}] when it is essential (e.g. \textit{substantial losses \underline{in cryospheric ecosystems}}).

\subsubsection{Relation Type}
\textbf{[\textit{Relation Type}] (\textit{positive}/\textit{negative})} describes the type of relation between cause and effect, i.e., \textit{positive} (cause/production/facilitation) or \textit{negative} (prevention) \citep{goldvarg2001naive,sloman2009causal}\footnote{Two other relations exist (\textit{allow} and \textit{allow not}), yet they are considered weak \citep{goldvarg2001naive}. Hence, our focus on stronger relations \textit{cause} and \textit{prevention}.}. A positive relation indicates that the existence of the cause \textit{leads to} the existence of the effect. A negative relation suggests that the existence of the cause \textit{prevents} the existence of the effect. For instance, a negative relation  between \textit{well-implemented land-based mitigation options} $\rightarrow$ \textit{trade-offs in terms of employment} can be triggered by the target \textit{``to avoid''}.

\subsubsection{Correlation} \label{sec:correlation}
\textbf{[\textit{Correlation}] (\textit{positive/negative})} marks the direction of the association between cause and effect. A correlation is classified as \textit{positive} when a change in the cause produces a change in the effect in the \textit{same} direction. It is \textit{negative} when the change occurs in the \textit{opposite} direction. Since cause and effect expressions in [\textit{Cause--NP}] and [\textit{Effect--NP}] frequently contain quantifiers or directional cues (e.g., \textit{``reductions of''} and \textit{``greater''}), the annotators manually remove these lexical signals to isolate the underlying concepts. These reformulations are stored in \textbf{[\textit{Cause--No\_Quantifier}] (\textit{string})} and \textbf{[\textit{Effect--No\_Quantifier}] (\textit{string})}. 
The value of [\textit{Correlation}] is then determined based on the relationship between the quantifier-free forms.

\subsection{Annotation Procedure}

We recruit two expert annotators with an academic background in linguistics and argumentation, who have been involved in a previous annotation campaign on causal discovery from text (more details in Appendix \ref{appendix:annotation_round_1}).

\subsubsection{Annotation Round 1}

The first annotation round starts with an oral discussion of the annotation guidelines, after which the two annotators independently annotate a subset of 11 statements (14.67\% of the dataset). Their annotations are compared, and disagreements are resolved by a third annotator.

There is perfect agreement on the presence and absence of causality (§\ref{sec:presence_absence_causality}): Cohen’s $\kappa = 1.0$; 10 \textit{yes}, 1 \textit{no}. This suggests that causality is clearly signaled in the statements and/or that the annotators share a consistent understanding of causality. In contrast, agreement on target identification (§\ref{sec:target_word_implicit}) is substantially lower, with $k = -.075$, indicating less-than-chance agreement. This low level of agreement extends to the other causal annotations. Implicit causal relations and nested causal structures, especially those involving report-specific abbreviations such as \textit{CO\textsubscript{2}-FFI} and \textit{CO\textsubscript{2}-LULUCF}, were frequently identified by one annotator only. Violations against multi-event decomposition were also common. In the subsequent feedback session with the expert annotators, it became clear that disagreements were most often due to gaps in domain knowledge (e.g., unfamiliarity with report-specific abbreviations) and vagueness in the annotation guidelines. Annotators typically accepted the causal relations identified by the other once clarified. We revised the annotation guidelines based on these observations.

\subsubsection{Annotation Round 2}

The second round adopts a validation-based approach to reduce the annotation time and cost. The more expressive annotator is asked to annotate all remaining statements. Here, \textit{expressiveness} is treated as a recall metric over the annotated causal relations in the first round; i.e., the expressiveness of annotator A is defined as the number of causal relations identified by annotator A and B, divided by the total number of causal relations annotated by B. The less expressive annotator then performs one of the following four actions on each causal relation identified and annotated by the first annotator, motivating the action in a free-text comment:
\begin{itemize}
    \item \textit{Valid}: causal relation appears in the statement and all annotations are correct. The annotations are kept as such.
    \item \textit{Remove}: causal relation does not appear in the statement. The annotations are therefore invalid and should be removed entirely.
    \item \textit{Add}: causal relation in the statement has not been identified by the first annotator. The second annotator provides the annotations. 
    \item \textit{Change}: causal relation appears in the statement, but not all annotations are correct. The second annotator changes the annotations and indicates what has been changed.
\end{itemize}

By adopting this kind of correction-based annotation setup, we aimed to reduce the cognitive load for the second annotator and specifically target significant errors (\textit{validate}/\textit{change}), omissions (\textit{add}), and redundancies (\textit{remove}).

The second annotator deemed all causal relation annotations valid for 43 statements and proposed actions for the remaining statements. The majority involved changes to the effect reformulations in [\textit{Effect--NP}], motivated by the redundant inclusion of the target word in the reformulation. In the end, we decided to keep the target words in the formulation of cause and/or effect when inclusion resulted in a more truthful representation of the event.

\section{Dataset Analysis}

\subsection{General Insights} Table \ref{tab:statistics} summarizes the dataset statistics. Causation seems to be a profound framing device for discussing climate change processes and mitigation strategies, with 63 out of 75 statements each reporting 14.06 causal relations on average. IPCC authors express causality primarily through positive relation types. Cause-effect associations are typically positive. Chi-squared tests indicate a significant association ($N=874, p<0.01$) between correlation type and explicitness ($\chi^2= 61.31$), relation type ($\chi^2= 86.14$), and nested causality ($\chi^2= 26.53$). Explicit relations occur more frequently than implicit ones, especially for negative correlations. Negative relation types are strongly linked to negative correlations, whereas nested relations exclusively exhibit positive correlations.

\begin{table}[t!]
    \centering
    \scriptsize
    \begin{tabular}{p{3.7cm} p{3cm}}
    \toprule
    \textbf{Feature} & \textbf{Count} \\
    \midrule
    Sections / paragraphs & 10 / 19 \\
    Statements & 75 \\
    \quad [\textit{Causation}]: \textcolor{blue}{Yes} $|$ \textcolor{red}{No} &
    \begin{tikzpicture}
        \draw[fill=blue!70] (0,0) rectangle (2.52,0.4); 
        \draw[fill=red!70] (2.52,0) rectangle (3.00,0.4); 
        \node[font=\tiny, white] at (1.26,0.2) {63};
        \node[font=\tiny, white] at (2.76,0.2) {12};
    \end{tikzpicture} \\
    Causal relations ; --NP & 874 \\
    \quad [\textit{Explicitness}]: \textcolor{blue}{Explicit} $|$ \textcolor{red}{Implicit} & 
    \begin{tikzpicture}
        \draw[fill=blue!70] (0,0) rectangle (2.04,0.4); 
        \draw[fill=red!70] (2.04,0) rectangle (3.00,0.4); 
        \node[font=\tiny, white] at (1.02,0.2) {593};
        \node[font=\tiny, white] at (2.52,0.2) {281};
    \end{tikzpicture} \\
    \quad [\textit{Relation type}]: \textcolor{blue}{Positive} $|$ \textcolor{red}{Negative} & 
    \begin{tikzpicture}
        \draw[fill=blue!70] (0,0) rectangle (2.835,0.4); 
        \draw[fill=red!70] (2.835,0) rectangle (3.00,0.4); 
        \node[font=\tiny, white] at (1.417,0.2) {827};
        \node[font=\tiny, white] at (2.917,0.2) {47};
    \end{tikzpicture} \\
    \quad [\textit{Correlation}]: \textcolor{blue}{Positive} $|$ \textcolor{red}{Negative} & 
    \begin{tikzpicture}
        \draw[fill=blue!70] (0,0) rectangle (2.01,0.4); 
        \draw[fill=red!70] (2.01,0) rectangle (3.00,0.4); 
        \node[font=\tiny, white] at (1.005,0.2) {581};
        \node[font=\tiny, white] at (2.505,0.2) {293};
    \end{tikzpicture} \\
    \quad [\textit{Nested}]: \textcolor{blue}{No} $|$ \textcolor{red}{Yes} & \begin{tikzpicture}
    \draw[fill=blue!70] (0,0) rectangle (2.84,0.4); 
    \draw[fill=red!70] (2.84,0) rectangle (3.00,0.4); 
    \node[font=\tiny, white] at (1.42,0.2) {828};
    \node[font=\tiny, white] at (2.92,0.2) {46};
\end{tikzpicture} \\
    \quad \textit{Overarching structures:} \textcolor{blue}{No} $|$ \textcolor{red}{Elaboration of examples} $|$ \textcolor{gray}{Combination of events} & \begin{tikzpicture}
    \draw[fill=blue!70] (0,0) rectangle (1.76,0.4); 
    \draw[fill=red!70] (1.76,0) rectangle (2.65,0.4); 
    \draw[fill=gray!70] (2.65,0) rectangle (3.00,0.4); 
    \node[font=\tiny, white] at (0.88,0.2) {517};
    \node[font=\tiny, white] at (2.205,0.2) {254};
    \node[font=\tiny, white] at (2.825,0.2) {103};
\end{tikzpicture} \\
    Unique relations; --No\_Quantifier & 653 \\
    Unique target words & 95 \\
    \bottomrule
    \end{tabular}
    \caption{General dataset statistics.}
    \label{tab:statistics}
\end{table}

\subsection{Readability and Semantic Complexity of Reported Causality}

The annotators reported a high cognitive load during causal relationship identification. They attributed this to the low readability of the statements in \textit{ClimateCause}. Readability is indeed low: most statements require college-level reading; Flesch Reading Ease $\in [0,30]$ \citep{flesch1948new}\footnote{More extensive readability analyses in Appendix \ref{appendix:readability_metrics}.}. However, traditional readability metrics, which are typically based on lexical complexity or sentence length, do not capture how easily readers can understand and retrieve reported causal relationships \citep{kincaid1975derivation,coleman1975computer,chall1995readability}. 

We therefore propose metrics to measure readability through the \textit{semantic complexity of the underlying causal structures}. Here, high complexity implies low readability. While structural properties such as graph depth and breadth mainly contribute to the structural complexity of a causal graph, we base our metrics on annotated causal properties such as explicitness and correlation. Relying on such semantic aspects go beyond existing metrics based on counts of causal connectives, verbs, and particles \citep{follmer2021predicting}. 

We assume that IPCC authors generally follow Grice’s maxim of manner, i.e., avoid unnecessary complexity \citep{grice1975}, as they want to inform non-experts about the current scientific consensus on climate change. Under this assumption, less frequent causal patterns are treated as more complex. The metrics therefore penalize causal graphs with common cause/effect structures, elaborations of examples, nested causality, and causal relations with negative correlation or relation type. We include additional penalties when these properties occur frequently and/or sequentially and encompass a large number of relations. 

\subsubsection{Readability Metrics}

\paragraph{Elaboration of examples and common cause/effect structures}  
Both structures include a set of events that have a subordinate relation to a shared overarching event, where the combination of events in the set is either facultative (i.e., \textit{elaboration of examples}) or mandatory (i.e., \textit{common cause/effect}) in the representation of causality. 

We identify the subordinating relations, i.e., event $\rightarrow$ overarching event, from [\textit{Cause--No\_Quantifier}] to [\textit{Cause--Belongs\_to}] and [\textit{Effect--No\_Quantifier}] to [\textit{Effect--Belongs\_to}] in $s$. Let $R^{com}$ be the set of relations with [\textit{Combined}] (\textit{yes}), and $R^{ex}$ those with [\textit{Combined}] (\textit{no}) and [\textit{Nested}] (\textit{no}). We construct graphs $G^{com}$ and $G^{ex}$ from those two sets, where nodes represent events and edges subordinating relations. We then decompose $G^{com}$ and $G^{ex}$ into connected subgraphs: $\hat{G}^{com} = \{\hat{G}^{com}_1,\dots,\hat{G}^{com}_{M}\}$ and $\hat{G}^{ex} = \{\hat{G}^{ex}_1,\dots,\hat{G}^{ex}_{N}\}$, where a subgraph $\hat{G}^{com}_m$ captures the $m^{\text{th}}$ overarching event with all its subordinating relations. Finally, complexity is computed as:
\begin{equation}
    C^{com}(s) = \sum_{i=1}^M\text{count\_edges}(\hat{G}^{com}_i) + M;
\end{equation}
\begin{equation}
    C^{ex}(s) = \sum_{i=1}^N \text{count\_edges}(\hat{G}^{ex}_i) + N;
\end{equation}
where the additive terms $M$ and $N$ penalize statements with many common cause/effect relations and elaboration of example structures.

\paragraph{Nested causality} 
These structures embed one or more causal relations in another event or concept. 

We associate all causal relations in $s$ for which [\textit{Nested}] (\textit{yes}) with its overarching ``nesting" event in [\textit{Effect--Belongs\_to}]. Let $L^{nest} = \{l_1,\dots,l_K\}$ be the set of nesting events and $T_k$ the number of causal relations that are nested within nesting event $l_k$. Complexity is computed as: 
\begin{equation}
    C^{nest}(s) = \sum_{i=1}^K(T_i+T_i\log T_i);
\end{equation}
where $T_i\log T_i$ penalizes events nesting a large number of relations, e.g., \textit{\underline{human}-caused climate change} (1 relation) vs \textit{CO\textsubscript{2}-LULUCF: carbon dioxide emissions from \underline{land use}, \underline{land-use change}, and \underline{forestry}} (3 relations).

\paragraph{Correlation} 
Negatively correlated causal events imply a change in the opposite direction when intervening on the cause. We construct a directed graph $G = (V,E)$ using all cause-effect pairs in $s$, where nodes $V$ represent events, edges $E$ represent causal relations, and each edge is labeled by the correlation between connected events. We compute $|E^-|$, the total number of negatively labeled edges; $|E^-_{path}|$, the number of paths containing two or more consecutive negative edges; $|V_{in}^{mix}|$, the number of nodes with both positive and negative \textit{incoming} edges; and $|V_{out}^{mix}|$, the number of nodes with both positive and negative \textit{outgoing} edges. Complexity is scored as: 
\begin{equation}
    C^{corr}(s) = |E^-|+|E^-_{path}|+|V_{in}^{mix}|+|V_{out}^{mix}|;
\end{equation}
penalizing graphs with a high frequency and long sequences of negatively correlated causal events, and a mix of positive and negative correlations.

\paragraph{Relation type} Complexity is computed analogously to $C^{corr}(s)$ with relation type annotations as edge labels: 
\begin{equation}
    C^{pol}(s) = |E^-|+|E^-_{path}|+|V_{in}^{mix}|+|V_{out}^{mix}|;
\end{equation}
penalizing graphs with a high frequency and long sequences of negative relation types, and a mix of positive and negative relation types.

\paragraph{Total semantic complexity of reported causality} The complexity metrics show distinct ranges in \textit{ClimateCause}: $C^{com} = [0,12]$, $C^{ex} = [0,16]$, $C^{nest} = [0,20.93]$, $C^{corr} = [0,70]$, and $C^{pol} = [0,40]$. Metrics with larger ranges may bias the overall complexity, $C$, when simply summed. We therefore normalize each metric $C^i(s)$, with $i \in \{com, ex, nest, corr, pol\}$ through min-max normalization based on its observed range across all statements, such that $\tilde{C}^{i}(s) \in [0,1]$: 
\begin{equation}
    \tilde{C}^{i}(s) = \frac{C^{i}(s) - \min(C^{i})}{\max(C^{i}) - \min(C^{i})}
\end{equation}
\begin{equation}
    C(s) = \sum^{5}_{i=1}w_i \tilde{C}^{i}(s);
\end{equation}
keeping $w_i$ constant ($=1$) in this work.

\subsubsection{Discussion}

In \textit{ClimateCause}, 43 statements (57.33\%) exhibit semantically complex causal structures, indicated by $C(s) > 0$ (max. $C(s) = 1.821$). Most graphs are complex in one (27 statements) or two (10 statements) metrics. None involves all five. 
We also observe a significant positive Pearson correlation between statement length (token count) and total complexity; $n=43, r = .590, p < 0.01$. This is expected since longer statements may include more causal relations, which potentially increases the semantic complexity of a causal graph. More analysis are included in Appendix \ref{appendix:readability_metrics_reported_causality}. 

While our metrics are frequency-driven and tailored to the semantic annotations in \textit{ClimateCause}, they lay the groundwork for topic-agnostic readability metrics that estimate cognitive complexity in interpreting causality in text. Future work could explore cognitive validation through user studies, adaptive weighting schemes, and integration with visualization tools to dynamically simplify complex causal structures for diverse audiences.

\section{Benchmarking Causal Reasoning}

\subsection{Problem Formalization} Causal reasoning is the ability to discern cause-and-effect relationships between variables from available data and draw causal inferences \citep{jin2023cladder,wang-2024-causalbench,yu-etal-2025-causaleval}. We argue that \textit{ClimateCause} is valuable for evaluating causal reasoning through multiple causality-specific tasks, such as causal event extraction, including event detection and event argument extraction \citep{simon-etal-2024-generative}, causal DAG inference \citep{kiciman2023causal}, and implicit causal chain discovery \citep{allein2025assessing}, but also through more general tasks like reading comprehension (RC). Here, we focus on correlation inference and causal chain reasoning.

\paragraph{Correlation Inference (CorrI)} CorrI is the ability to identify the direction of association between two variables, where positive correlation means that the variables change in the \textit{same} direction and negative the \textit{opposite} direction. \begin{itemize}
    \item \textbf{CorrI}: Given a set of causal pairs $R$;
    \item \textbf{CorrI+RC}: Given a statement $s$ and a set of causal pairs $R$.
\end{itemize}
For each $(e_i,e_j) \in R$, determine whether the correlation between $e_i$ and $e_j$ is positive or negative. Label distribution is \{\textit{positive}: 581, \textit{negative}: 293\}.

\paragraph{Causal Chain Reasoning (CCR)} CCR is the ability to identify and analyze causal chains, where a causal chain is defined as a directed path of at least three nodes in a causal graph \citep{pearl2009causality}. 
\begin{itemize}
    \item \textbf{CCR}: Given a causal graph where nodes $V$ represent events, and edges causal relations.
    \item \textbf{CCR+ECI+RC}: Given a statement $s$ and the set of all causal events in $s$, $V$.
\end{itemize}
In CCR+ECI+RC, the causal relations between the events in $V$ need to be inferred from the text (i.e., event causality inference (ECI)) before a model can reason over their membership or position in a causal graph. \textbf{\textit{Membership}}: Determine for $e \in V$ whether it is part of a causal chain structure; \{\textit{yes}: 115, \textit{no}: 397\}. \textbf{\textit{Position}}: Determine for $e \in V$ which position it holds in a causal chain; \{\textit{start} (= source node): 32, \textit{middle} (= mediator node): 48, \textit{end} (= sink node): 35, \textit{none}: 397\}. 

\begin{table}[t]
    \centering
    \footnotesize
    \begin{subtable}[t]{0.48\textwidth}
    \centering
    \begin{tabular}{p{1.2cm}|p{1.6cm}p{1.6cm}p{1.6cm}}
        \toprule
        \textbf{Prompt} & \textbf{Precision} & \textbf{Recall} & \textbf{F1} \\
        \midrule
        \multicolumn{4}{c}{\textbf{CorrI}} \\
        \midrule
        0-shot & $0.8204_{\pm 0.1204}$ & $0.8003_{\pm 0.2032}$ & $0.7887_{\pm 0.0738}$ \\
        F-shot & $0.8874_{\pm 0.1017}$ & $0.9486_{\pm 0.0211}$ & $0.9156_{\pm 0.0652}$ \\
        CoT & $0.8556_{\pm 0.0971}$ & $0.9369_{\pm 0.0364}$ & $0.8934_{\pm 0.0683}$ \\
        \rowcolor{gray!20}\textit{Avg} & $0.8544_{\pm 0.0970}$ & $0.8953_{\pm 0.1259}$ & $0.8659_{\pm 0.0839}$ \\
        \midrule
        \multicolumn{4}{c}{\textbf{CorrI+RC}} \\
        \midrule
        0-shot & $0.8294_{\pm 0.1330}$ & $0.7194_{\pm 0.1294}$ & $0.7574_{\pm 0.0391}$ \\
        F-shot & $0.9426_{\pm 0.0644}$ & $0.9641_{\pm 0.0209}$ & $0.9529_{\pm 0.0430}$ \\
        CoT & $0.8621_{\pm 0.0858}$ & $0.8789_{\pm 0.0820}$ & $0.8678_{\pm 0.0635}$ \\
        \rowcolor{gray!20}\textit{Avg} & $0.8780_{\pm 0.0993}$ & $0.8542_{\pm 0.1325}$ & $0.8594_{\pm 0.0952}$ \\
        \bottomrule
    \end{tabular}
    \caption{Correlation inference.}
\end{subtable}
    \hfill
    \begin{subtable}[t]{0.48\textwidth}
    \centering
    \begin{tabular}{p{1.2cm}|p{1.6cm}p{1.6cm}p{1.6cm}}
    \toprule
    \textbf{Prompt} & \textbf{Precision} & \textbf{Recall} & \textbf{F1} \\
    \midrule
    \multicolumn{4}{c}{\textbf{CCR \textit{membership}}} \\
    \midrule
    Adjacency & $0.5382_{\pm 0.3205}$ & $0.8754_{\pm 0.0329}$ & $0.6308_{\pm 0.2192}$ \\
    Single N. & $0.5023_{\pm 0.2619}$ & $0.9739_{\pm 0.0151}$ & $0.6371_{\pm 0.2087}$ \\
    GraphML & $0.6421_{\pm 0.3130}$ & $0.9362_{\pm 0.1030}$ & $0.7205_{\pm 0.1601}$ \\
    \rowcolor{gray!20}\textit{Avg} & $0.5609_{\pm 0.2670}$ & $0.9285_{\pm 0.0695}$ & $0.6628_{\pm 0.1766}$ \\
    \midrule
    \multicolumn{4}{c}{\textbf{CCR \textit{position}}} \\
    \midrule
    Adjacency & $0.2927_{\pm 0.0757}$ & $0.9253_{\pm 0.0686}$ & $0.4417_{\pm 0.0893}$ \\
    Single N. & $0.2914_{\pm 0.0878}$ & $0.9615_{\pm 0.0666}$ & $0.4432_{\pm 0.1043}$ \\
    GraphML & $0.2667_{\pm 0.0646}$ & $0.9885_{\pm 0.0199}$ & $0.4175_{\pm 0.0791}$ \\
    \rowcolor{gray!20}\textit{Avg} & $0.2836_{\pm 0.0676}$ & $0.9585_{\pm 0.0560}$ & $0.4341_{\pm 0.0802}$ \\
    \midrule
    \multicolumn{4}{c}{\textbf{CCR+ECI+RC \textit{membership}}} \\
    \midrule
    0-shot & $0.2860_{\pm 0.0080}$ & $0.9014_{\pm 0.0411}$ & $0.4339_{\pm 0.0047}$ \\
    F-shot & $0.3661_{\pm 0.1218}$ & $0.6899_{\pm 0.2016}$ & $0.4514_{\pm 0.0482}$ \\
    CoT & $0.3532_{\pm 0.1132}$ & $0.7855_{\pm 0.1267}$ & $0.4718_{\pm 0.0728}$ \\
    \rowcolor{gray!20}\textit{Avg} & $0.3351_{\pm 0.0912}$ & $0.7923_{\pm 0.1517}$ & $0.4524_{\pm 0.0467}$ \\
    \midrule
    \multicolumn{4}{c}{\textbf{CCR+ECI+RC \textit{position}}} \\
    \midrule
    0-shot & $0.1851_{\pm 0.0160}$ & $0.6706_{\pm 0.1336}$ & $0.2896_{\pm 0.0324}$ \\
    F-shot & $0.1980_{\pm 0.0823}$ & $0.5545_{\pm 0.1170}$ & $0.2781_{\pm 0.0650}$ \\
    CoT & $0.2600_{\pm 0.0641}$ & $0.5087_{\pm 0.2241}$ & $0.3405_{\pm 0.1094}$ \\
    \rowcolor{gray!20}\textit{Avg} & $0.2144_{\pm 0.0631}$ & $0.5780_{\pm 0.1602}$ & $0.3027_{\pm 0.0717}$ \\
    \bottomrule
    \end{tabular}
    \caption{Causal chain reasoning.}
\end{subtable}
    \caption{Mean and standard deviation results for each prompting strategy individually (3 runs) and all strategies together (9 runs; \textit{Avg}).}
    \label{tab:benchmarking_results}
\end{table}

\subsection{Methodology} 

We assess GPT5.1 through in-context learning. CorrI, CorrI+RC, and CCR+ECI+RC are evaluated through (i) zero-shot, (ii) few-shot, and (iii) chain-of-thought prompting. For CCR, we select three prompting strategies defined in \citet{sheth-etal-2025-causalgraph2llm}, where the encoding of the given causal graph (i) lists all edges (adjacency), (ii) textually presents each node with its direct effects (single node), or (iii) uses GraphML format. Each prompting strategy is tested using three prompt variations to enforce robustness of the results. Events are represented using [\textit{Cause-No\_Quantifier}] and [\textit{Effect-No\_Quantifier}]. Table \ref{tab:benchmarking_results} shows the results, with per-class performance for \textit{\textbf{position}} in Appendix \ref{appendix:benchmarking}.

\subsection{Analysis}

\paragraph{Correlation inference} The results for CorrI and CorrI+RC are both high and comparable. This suggests that the LLM effectively captures the correlation between events and that the presence of the original statement, which often verbalizes the correlation through adjectives and verbs, has little impact on prediction performance. We verify this by comparing the predictions with (CorrI+RC) and without (CorrI) the statement using the McNemar test \citep{mcnemar1947note}, which focuses on discordant results. A Bonferroni correction ($\alpha = 0.05/9 \approx 0.0056$) was applied to control Type I errors \citep{bonferroni1936teoria}. The results show significant discordance in five prompt variations. This indicates that statement availability does affect predictions, though not improving the performance.

\paragraph{Causal chain inference} 

Despite promising recall performance, the LLM does not seem to fully understand what causal chains are or consistently infer them as part of its reasoning. For example, one main feature of a chain is that it involves at least three events. A manual error analysis of the membership predictions shows that for 45 statements without chains, the LLM (in at least one of its prompt settings) predicts only one or two events as chain members in 37 cases (CCR+ECI+RC). This may be due to the abstract causal graph-oriented definition of a causal chain in the prompt, which does not align well with text-only input. However, this behavior persists in 23 statements when the input includes an explicit causal graph (CCR). 

Chain membership identification is an inherent step prior to position identification. If the LLM takes this step during position identification, events outside a chain (\textit{membership = no}) should have \textit{position = none}, and those within (\textit{membership = yes}) should have \textit{position =} \textit{start}, \textit{middle}, or \textit{end}. However, McNemar tests reveal significant discordance in 8 (CCR) and 7 (CCR+ECI+RC) of 9 prompt pairs. The LLM also frequently over-predicts start, middle, and end positions, with low mean precision scores: $0.2687/0.5008/0.3123$ (CCR) and $0.1934/0.2788/0.2283$ (CCR+ECI+RC). Confusion matrices show that most false positives belong to the `none' class, followed by the middle class for start and end instances. Start and end are rarely confused. 

Last, our causality-focused readability metric, $C(s)$, shows meaningful variation across predicted membership and position labels in CCR+ECI+RC. Kruskal-Wallis H tests \citep{kruskal1952use} indicate significant differences in $C(s)$ across the predicted labels ($p<0.05$). For \textit{\textbf{membership}}, all nine runs are significant: $H(1) \in [9.55, 84.70]$, with small to large effect sizes ($\epsilon^2 \in [0.02,0.16]$). For \textit{\textbf{position}}, five runs are significant: $H(3) \in [7.92, 47.01], p<0.05$, with small to moderate effect sizes, $\epsilon^2 \in [0.01,0.09]$. Dunn’s post‑hoc comparisons \citep{dunn1964multiple} with Holm correction reveal no significant pairwise differences, suggesting subtle distributed effects rather than strong contrasts between labels.

\section{Conclusion}

This work introduced a manually annotated dataset designed to uncover complex causal structures in climate change reports, with unique annotations for correlation, nested causality, and spatiotemporal contextualization. Beyond its value for benchmarking correlation inference and causal chain reasoning in large language models, we demonstrated its potential for broader applications such as causality-focused readability metrics. Looking ahead, we consider spatiotemporal contextualization and scientific confidence assessments of causal relations particularly interesting avenues for advancing confounder and causal strength inference.

\section*{Limitations}

Properties of causal relations not specifically addressed in this annotation setup but that are potentially relevant in the climate change discussion include causal strength and causal uncertainty. Parallels could be drawn with the confidence level labels that the report writers assigned to the statement. However, it is unclear whether the confidence level applies to every causal relation in that statement as the confidence evaluation does not focus solely on causality. \citet{molina2021evolution}, for example, discovered a shift over time towards higher certainty levels in IPCC reports, implying a “call to action”. These properties were excluded due to the difficulty of reliably inferring them from text without additional context or expert judgment.

The readability metrics are based on the semantic complexity of reported causal structures, where they penalize properties that occur less frequent in the \textit{ClimateCause} dataset, adding additional penalities for structures that include a high number of such properties. Aligning these metrics with findings in the cognitive science and examining the frequencies of the observed causal patterns in \textit{ClimateCause} would provide stronger, dataset-agnostic motivations for the metrics. Moreover, while the weights in complexity metric $C(s)$ are kept constant, one can arguably look into learning these weights empirically or normalizing them based on the observed ranges.

\textit{ClimateCause} is moderate in size and heavily focused on a single topic. The benchmarking results on correlation inference and causal chain reasoning should not be seen as a generalizable evaluation of LLM capabilities. Larger and topic-diverse resources are necessary to make generalizable conclusions. Since \textit{ClimateCause} is a newly constructed and does not modify or adapt an existing causal reasoning benchmark, it avoids the issues and limitations related to benchmarking on reused datasets, such as data contamination in the pre-training data of the LLM \citep{bean2025measuring}. 

\section*{Ethical Considerations}

Our use of the data from the Semantic Climate Wikibase is consistent with the intended use described in the Apache License 2.0. All annotations were made voluntarily and without remuneration. Given the nature of the IPCC data, the annotated content is free of any material that could negatively affect the integrity of the annotators or their personal well-being.

\section*{Acknowledgements} 

This work was funded by the Research Foundation - Flanders (FWO) under grant G0L0822N and the Swiss National Science Foundation (SNSF) under grant 209674 through the CHIST-ERA project ``iTRUST Interventions against Polarisation in Society for Trustworthy Social Media. From Diagnosis to Therapy''. Liesbeth Allein is further supported by a junior postdoctoral fellowship from the FWO under grant 12AGW26N.

\bibliography{aaai2026}

\appendix
\section{IPCC Sections} \label{appendix:ipcc_sections}

Table \ref{tab:ipcc-structure} displays a breakdown of all sections of the IPCC Climate Change 2023 Report from which the statements in \textit{ClimateCause} are taken.

\begin{table}[htbp!]
\centering
\scriptsize
\begin{tabular}{p{0.8cm} p{6cm}}
\hline
\textbf{Section} & \textbf{Title} \\
\hline
\textbf{2} & Current Status and Trends \\[0.3em]
2.1 & Observed Changes, Impacts and Attribution \\[0.3em]
2.1.1 & \textit{Observed Warming and its Causes} \\[0.3em]
2.1.2 & \textit{Observed Climate System Changes and Impacts to Date} \\[0.3em]
2.3 & Current Mitigation and Adaptation Actions and Policies are not Sufficient \\[0.3em]
2.3.2 & \textit{Adaptation Gaps and Barriers} \\[0.3em]
\textbf{3} & Long-Term Climate Change and Development Futures \\[0.3em]
3.1 & Long-Term Climate Change, Impacts and Related Risks \\[0.3em]
3.1.1 & \textit{Long-term Climate Change} \\[0.3em]
3.1.3 & \textit{The Likelihood and Risks of Abrupt and Irreversible Change} \\[0.3em]
\textbf{4} & Near-Term Responses in a Changing Climate \\[0.3em]
4.1 & \textit{The Timing and Urgency of Climate Action} \\[0.3em]
4.8 & Strengthening the Response: Finance, International Cooperation and Technology \\[0.3em]
4.8.1 & \textit{Finance for Mitigation and Adaptation Actions} \\[0.3em]
4.9 & \textit{Integration of Near-Term Actions Across Sectors and Systems} \\
\hline
\end{tabular}
\caption{Sections of the IPCC Climate Change 2023 Report from which the texts in \textit{ClimateCause} have been drawn.}
\label{tab:ipcc-structure}
\end{table}

\section{Annotation Guidelines} \label{appendix:annotation_guidelines}

We suggested to the annotators to follow a three-step annotation flow when identifying and annotating causal relations for a given statement:
\begin{enumerate}
    \item \textbf{Causal relation extraction:} Decide whether the statement expresses at least one causal relation. If so, identify all the cause-effect pairs, specify their context mentioned in the statement, highlight the terms in the statement that trigger the causal relation between the events or actions that make up the cause and effect, and indicate whether the causality is explicitly or implicitly conveyed;
    \item \textbf{Standardization and characterization:} Standardize the phrasing of the cause and effect by formulating the events into noun phrases and characterize the relation type and correlation of the causal relation;
    \item \textbf{Complex causal structures:} Identify and label causal structures present in the statement, such as common cause/effect.
\end{enumerate}

\subsection{Annotation Setting}

The annotators were presented with statements from the IPCC reports, where each statement (ranging from single sentences to full paragraphs) was shown on a line in an excel file. The features they had to annotate were assigned their own column, which they had to fill out.

\subsection{Annotation Features}

Table \ref{app:overview-features} gives an overview of the features annotated in \textit{ClimateCause}.
\begin{table}[ht]
\centering
\scriptsize
\begin{tabular}{|p{2.3cm}|p{4.5cm}|}
\hline
\textbf{Feature} & \textbf{Description} \\
\hline
Statement link & URL to the statement in Wikibase (no action needed) \\
Section & Section number from which the statement is taken (no action needed) \\
Paragraph & Paragraph number from which the statement is taken (no action needed) \\
Series ordinal & Position in the paragraph from which the statement is taken (no action needed) \\
Confidence level & Confidence level of the statement (no action needed) \\
Statement & The statement (no action needed) \\
\textit{Causation} & Binary indicator (yes/no) whether the statement reports a causal relation (§\ref{causation}) \\
\textit{Target} & Target word(s) (string) that evokes the causal relation (§\ref{target}) \\
\textit{Cause -- NP} & Noun phrase reformulation of the cause (string) (§\ref{np-formulation}) \\
\textit{Cause -- Context} & Spatiotemporal context of the cause (string) (§\ref{app:modspacetime}) \\
\textit{Cause -- No\_quantifier} & Reformulation of the cause without quantifiers (string) (§\ref{no-quantifiers}) \\
\textit{Cause -- Belongs\_to} & Event to which the cause belongs (string) (§\ref{sec:multi-component-events}) \\
\textit{Effect -- NP} & Noun phrase reformulation of the effect (string) (§\ref{np-formulation}) \\
\textit{Effect -- Context} & Spatiotemporal context of the effect (string) (§\ref{app:modspacetime}) \\
\textit{Effect -- No\_quantifier} & Reformulation of the effect without quantifiers (string) (§\ref{no-quantifiers}) \\
\textit{Effect -- Belongs\_to} & Event to which the effect belongs (string) (§\ref{sec:multi-component-events}) \\
\textit{Combined} & Binary indicator (yes/no) whether the connection between the cause/effect in \textit{--NP} and the overarching event in \textit{--Belongs\_to} is binding \\
\textit{Nested causality} & Binary indicator (yes/no) whether the causal relation is the nested part in a nesting construction \\
\textit{Explicitness} & Binary label (E/I) whether the causal relation is conveyed explicitly or implicitly (§\ref{explicitness}) \\
\textit{Relation type} & Binary label (positive/negative) whether the relation type is positive (CAUSES) or negative (PREVENTS) (§\ref{relation type}) \\
\textit{Correlation} & Binary label (positive/negative) whether correlation is positive (increase $\to$ increase) or negative (increase $\to$ decrease) (§\ref{correlation}) \\
\textit{Abbreviations} & Set of abbreviations used in the statement resolved to their full meaning (string) (§\ref{abbreviations}) \\
\hline
\end{tabular}
\caption{Overview of all features with their description.}
\label{app:overview-features}
\end{table}

\subsection{Presence/Absence of Causal Relations [\textit{Causation}]}\label{causation}

Given a statement, indicate whether or not it contains at least one causal relation [\textit{Causation}]. 
\begin{itemize}
    \item If no causal relation is present: [\textit{Causation}] = No. 
\end{itemize}
In this case, no further annotation action needs to be taken for the statement.
\begin{itemize}
    \item If at least one causal relation is present: [\textit{Causation}] = Yes. 
\end{itemize}

\subsection{Target Identification [\textit{Target}]} \label{target}

Write down the [\textit{Target}] that evokes the causal relation. 

\paragraph{Span of the target} When the target is a \textbf{verb}, include:
\begin{itemize}
    \item All `verb' parts of the verb (e.g., has caused, is causing, can cause, will continue to increase, cannot avoid).
    \item Prepositions in case of preposition-combined verbs (e.g., arise from, contribute to, lead to, account for).
    \item Words that would otherwise break the span (e.g., driven in part by, have adversely affected, can also cause).
    \item `By' in passive constructions (e.g., can be avoided by, exacerbated by).
    \item `To' for infinitives (e.g., to minimize, to address).
\end{itemize}
If target is a \textbf{noun} (rather rare), include:
\begin{itemize}
    \item Determiner/article `a' and `the' (e.g., a cause of, the effect of).
    \item Adjectives and quantifiers (e.g., the direct cause of, other causes of, historical and unequal contributions arising from).
\end{itemize}
If target is \textbf{other}, include:
\begin{itemize}
    \item If causality embedded in word, abbreviation, or adjective: give that word, abbreviation, or adjective as target (e.g., AFOLU, anthropogenic).
    \item Discourse markers (e.g., due to, from).
    \item Special case: ``human-caused climate change'': here, the target is ``caused'' with causal relation: humans (cause) $\rightarrow$ climate change (effect).
    \item If the target is not explicit, which means that the causal relation is indirect without explicit target, use ``/'' to mark the target.
\end{itemize}

\subsection{Identification of Causal Relations}

\subsubsection{General Rules}

\paragraph{Each entry presents one causal relation} Each causal relation in the statement is identified and assigned to a new annotation entry, copying the statement and all other non-modifiable metadata (i.e., statement link, section, paragraph, series ordinal, and confidence level) to a new row in the annotation file. This way, one row in the file points to one causal relation.

The cause is contained in [\textit{Cause--NP}] and the effect in [\textit{Effect--NP}], with \textit{--NP} referring to noun phrase reformulation (noun phrase reformulation is described in §\ref{np-formulation}).

\paragraph{Note on confidence level} A statement may mention multiple confidence levels.
Make sure to assign each confidence level to the appropriate causal relation.

\paragraph{Reference resolution} References such as \textit{``it''} are replaced with their referent, for which the annotators have access to the full text of the IPCC reports.

\subsubsection{Noun Phrase Reformulation \textit{(--NP)}} \label{np-formulation}

\paragraph{General rule} The cause and effect should be formulated as noun phrases in such a way that one can read the causal relation as \textit{``[Cause--NP] causes/leads to [Effect--NP]''}, and the causal relation can be understood as outside its communicative context. 

\paragraph{Reformulation guidelines} Make minimal alterations to the semantics of the cause and effect so that the reformulated version stays as close as possible to the semantics of the original phrasing. Rely on words used in the statement for reformulation as much as possible.
\\ \\
\textit{\textbf{Example:} ``With further global warming, every region is projected to increasingly experience concurrent and multiple changes in climatic impact-drivers.''}
\\ \\
The two causal relations are (\ref{app:sample-1}):
\begin{table}[ht!]
\centering
\small
\begin{tabular}{|p{3cm}|p{3.78cm}|}
\hline
\textbf{Cause--NP} & \textbf{Effect--NP} \\ \hline
further global warming & increase in concurrent changes in climatic impact-drivers \\ \hline
further global warming & increase in multiple changes in climatic impact-drivers \\ \hline
\end{tabular}
\caption{Example of [Cause--NP] and [Effect--NP].}
\label{app:sample-1}
\end{table}
\\ Here, \textit{``increase in''} is inspired by \textit{``increasingly experience''}. Moreover, the formulation of the effect as \textit{``increase in concurrent changes in climatic impact-drivers''} is more precise and closer to the effect report in the statement than \textit{``concurrent changes in climatic impact-drivers''}.
\\ \\
\textit{\textbf{Example:} ``At 2°C or above, [...] more frequent and/or severe agricultural and ecological droughts are projected in Europe, Africa, Australasia and North, Central and South America.''}
\\ \\
Four causal relations can be formulated as (\ref{app:sample-2}):
\begin{table}[t!]
\centering
\small
\begin{tabular}{|p{3cm}|p{3.78cm}|}
\hline
\textbf{Cause--NP} & \textbf{Effect--NP} \\ \hline
global warming of 2°C or above & higher frequency of agricultural droughts \\ \hline
global warming of 2°C or above & higher severity of agricultural droughts \\ \hline
global warming of 2°C or above & higher frequency of ecological droughts \\ \hline
global warming of 2°C or above & higher severity of ecological droughts \\ \hline
\end{tabular}
\caption{Example of [Cause--NP] and [Effect--NP].}
\label{app:sample-2}
\end{table}

Here, the cause \textit{``2°C or above''} is reformulated as \textit{``global warming of 2°C or above''} such that the cause can be understood outside of its context. To resolve this, annotators have access to the original texs of the IPCC reports.

\subsubsection{Multi-Component Events \textit{(--Belongs\_to)} [\textit{Combined)}]} \label{sec:multi-component-events}
Causal relations with the multi-component cause or effect are broken down into their individual causal relations, and each causal relation is assigned to its own line in the excel file. 
\paragraph{In general} Causes and effects are split into separate events.
\\ \\
\textit{\textbf{Example:} ``With further global warming, every region is projected to increasingly experience concurrent and multiple changes in climatic impact-drivers.''}
\\ \\
The two causal relations are (\ref{app:sample-3}):
\begin{table}[ht!]
\centering
\small
\begin{tabular}{|p{3cm}|p{3.78cm}|}
\hline
\textbf{Cause--NP} & \textbf{Effect--NP} \\ \hline
further global warming & increase in concurrent changes in climatic impact-drivers \\ \hline
further global warming & increase in multiple changes in climatic impact-drivers \\ \hline
\end{tabular}
\caption{Example of [Cause--NP] and [Effect--NP].}
\label{app:sample-3}
\end{table}

\paragraph{When multi-component event presents examples of an overarching event} The components can be presented as examples of an overarching event or concept. If the examples act as \textit{cause}, then the overarching event to which they belong is marked in [\textit{Cause--Belongs\_to}]. If the examples act as \textit{effect}, then the overarching event to which they belong is marked in [\textit{Effect--Belongs\_to}]
\\ \\
\textit{\textbf{Example:} ``Impacts in ecosystems from slow-onset processes such as ocean acidification, sea level rise or regional decreases in precipitation have also been attributed to human-caused climate change.''}
\\ \\
In the above case, there should be four lines in the excel sheet for each causal relation (\ref{app:sample-4}):
\begin{table}[ht!]
\centering
\small
\begin{tabular}{|p{1.7cm}|p{1.7cm}|p{1.7cm}|p{0.6cm}|}
\hline
\textbf{Cause--NP} & \textbf{Effect--NP} & \textbf{Cause--Belongs\_To} & \textbf{Com-bined} \\ \hline
slow-onset processes & impacts in ecosystems & -- & -- \\ \hline
ocean acidification & impacts in ecosystems & slow-onset processes & No \\ \hline
sea level rise & impacts in ecosystems & slow-onset processes & No \\ \hline
regional decreases in precipitation & impacts in ecosystems & slow-onset processes & No \\ \hline
\end{tabular}
\caption{Example of multi-component event with elaboration of examples of an overarching event.}
\label{app:sample-4}
\end{table}

\paragraph{On [\textit{Combined}]} In this case, [\textit{Combined}] = \textit{No} to indicate that the combination of the events in cause or effect is not a necessary condition for the validity of the causal relation.

\paragraph{When multi-component event presents a combination of events} The next example statement shows an instance where the causal relation is only established through the combination of the events in the cause. If the events in the combination act as \textit{cause}, then the overarching event to which they belong is marked in [\textit{Cause--Belongs\_to}]. If the events in the combination act as \textit{effect}, then the overarching event to which they belong is marked in [\textit{Effect--Belongs\_to}]
\\ \\
\textit{\textbf{Example:} ``Nearly 50\% of coastal wetlands have been lost over the last 100 years, as a result of the combined effects of localised human pressures, sea level rise, warming and extreme climate events.''}
\\ \\
In the above case, there should be five lines in the excel sheet for each causal relation (\ref{app:sample-5}):
\begin{table}[ht!]
\centering
\small
\begin{tabular}{|p{1.5cm}|p{1.5cm}|p{2.0cm}|p{0.8cm}|}
\hline
\textbf{Cause--NP} & \textbf{Effect--NP} & \textbf{Cause--Belongs\_To} & \textbf{Com-bined} \\ \hline
combined effects of localised human pressures, sea level rise, warming and extreme climate events & loss of nearly 50\% of coastal wetlands & -- & -- \\ \hline
localised human pressures & loss of nearly 50\% of coastal wetlands & combined effects of localised human pressures, sea level rise, warming and extreme climate events & Yes \\ \hline
sea level rise & loss of nearly 50\% of coastal wetlands & combined effects of localised human pressures, sea level rise, warming and extreme climate events & Yes \\ \hline
warming & loss of nearly 50\% of coastal wetlands & combined effects of localised human pressures, sea level rise, warming and extreme climate events & Yes \\ \hline
extreme climate events & loss of nearly 50\% of coastal wetlands & combined effects of localised human pressures, sea level rise, warming and extreme climate events & Yes \\ \hline
\end{tabular}
\caption{Example of multi-component event with combination of events.}
\label{app:sample-5}
\end{table}

\paragraph{On [\textit{Combined}]} Here, [\textit{Combined}] = \textit{Yes} to indicate that the causal relation is reported to be only established through the combination of events.

\subsubsection{Modifiers of Space and Time \textit{(--Context)}} \label{app:modspacetime}
Modifiers regarding space and time are often included in cause and effect. This spatiotemporal context of the cause and the context of the effect are annotated as [\textit{Cause--Context}] and [\textit{Effect--Context}], respectively. 
\\ \\
The basic rule is:

\textbf{$\rightarrow$ If the cause or effect is clear on itself and the modifier is \underline{NOT} necessary to understand the event.}
\\ \\
The cause/effect should be written down \textbf{without} the modifier and the modifier should be included in either [\textit{Cause--Context}] (when the event is a cause) or [\textit{Effect--Context}] (when the event is an effect). 
\\ \\
\textbf{\textit{Example:}} \textit{``Ocean warming and ocean acidification have adversely affected food production from shellfish aquaculture and fisheries in some oceanic regions (high confidence).''}:
\begin{itemize}
    \item \textit{``in some oceanic regions''} is not necessary to understand the effect \textit{decrease in food production from shellfish aquaculture} or the effect \textit{decrease in food production from fisheries};
    \item Therefore, [\textit{Effect--Context}] = \textit{``in some oceanic regions''}.
\end{itemize}

\textbf{$\rightarrow$ If the cause or effect is \underline{NOT} clear on itself and the modifier is necessary to understand the event:}
\\ \\
The cause/effect should be written down \textbf{together with} the modifier. 
\\ \\
\textbf{\textit{Example:}} \textit{``Climate change has caused substantial damages, and increasingly irreversible losses, in terrestrial, freshwater, cryospheric and coastal and open ocean ecosystems.''}:
\begin{itemize}
    \item \textit{``in terrestrial ecosystems} is necessary to capture the event more exactly;
    \item Therefore, [\textit{Effect--NP}] = \textit{``substantial damages in terrestrial ecosystems''}, [\textit{Effect--Context}] is empty.
\end{itemize}

\paragraph{In case of multiple modifiers} Some statements mention more than one spatiotemporal modifier for one causal relation. In this case, a distinction between \textit{necessary} and \textit{contextual} modifiers should be made. As with cause and effect, the modifiers should be split into single locations and times. The causal relation should then be repeated multiple times, each time annotated with one modifier, either as part of the cause/effect or as context in [\textit{Cause--Context}] or [\textit{Effect--Context}].
\\ \\
\textbf{\textit{Example:}} \textit{``Climate change has caused substantial damages, and increasingly irreversible losses, in terrestrial, freshwater, cryospheric and coastal and open ocean ecosystems.''}
\\ \\
In the above case, the modifiers are \textbf{necessary}. There should be ten lines in the excel sheet reflecting each modifier (\ref{app:sample-6}):
\begin{table}[ht!]
\centering
\small
\begin{tabular}{|p{2.5cm}|p{4.3cm}|}
\hline
\textbf{Cause--NP} & \textbf{Effect--NP} \\ \hline
climate change & substantial damages in terrestrial ecosystems \\ \hline
climate change & substantial damages in freshwater ecosystems \\ \hline
climate change & substantial damages in cryospheric ecosystems \\ \hline
climate change & substantial damages in coastal ecosystems \\ \hline
climate change & substantial damages in open ocean ecosystems \\ \hline
climate change & increasingly irreversible losses in terrestrial ecosystems \\ \hline
climate change & increasingly irreversible losses in freshwater ecosystems \\ \hline
climate change & increasingly irreversible losses in cryospheric ecosystems \\ \hline
climate change & increasingly irreversible losses in coastal ecosystems \\ \hline
climate change & increasingly irreversible losses in open ocean ecosystems \\ \hline
\end{tabular}
\caption{Example of spatiotemporal contextualization - \textbf{necessary} modifiers.}
\label{app:sample-6}
\end{table}
\\ \\
\textbf{\textit{Example:}} \textit{``Climate change has contributed to desertification and exacerbated land degradation, particularly in low lying coastal areas, river deltas, drylands and in permafrost areas.''}
\\ \\
In the above case, the modifiers are \textbf{contextual}. There should be ten lines in the excel sheet (\ref{app:sample-7}):
\begin{table}[ht!]
\centering
\small
\begin{tabular}{|p{2.1cm}|p{2cm}|p{2.1cm}|}
\hline
\textbf{Cause--NP} & \textbf{Effect--NP} & \textbf{Effect--Context} \\ \hline
climate change & desertification &  \\ \hline
climate change & exacerbated land degradation &  \\ \hline
climate change & desertification & in low lying coastal areas \\ \hline
climate change & exacerbated land degradation & in low lying coastal areas \\ \hline
climate change & desertification & in river deltas \\ \hline
climate change & exacerbated land degradation & in river deltas \\ \hline
climate change & desertification & in drylands \\ \hline
climate change & exacerbated land degradation & in drylands \\ \hline
climate change & desertification & in permafrost areas \\ \hline
climate change & exacerbated land degradation & in permafrost areas \\ \hline
\end{tabular}
\caption{Example of spatiotemporal contextualization - \textbf{contextual} modifiers.}
\label{app:sample-7}
\end{table}
\\ \\
The first two causal relations also are valid because of the word \textit{``particularly''}. It signals that the causal relation is particularly noticeable in the areas mentioned after it, but it does not exclude other areas. 

\paragraph{Alignment with confidence label} Contextual modifiers can also be linked to different confidence labels. In this case, make sure that the confidence label, which is given beforehand, aligns with the annotated causal relation.
\\ \\
\textit{\textbf{Example:} ``At 1.5°C global warming, heavy precipitation and flooding events are projected to intensify and become more frequent in most regions in Africa, Asia (high confidence), North America (medium to high confidence) and Europe (medium confidence).''}
\\ \\
Three of the sixteen causal relations that should be annotated look as follows (\ref{app:sample-8}):
\begin{table}[ht!]
\centering
\small
\begin{tabular}{|p{1.5cm}|p{1.5cm}|p{1.5cm}|p{1.5cm}|}
\hline
\textbf{Cause--NP} & \textbf{Effect--NP} & \textbf{Effect--Context} & \textbf{Confidence Level} \\ \hline
global warming of 1.5°C & higher intensity of heavy precipitation events & in most regions in Africa & high confidence \\ \hline
global warming of 1.5°C & higher intensity of heavy precipitation events & in most regions in North America & medium to high confidence \\ \hline
global warming of 1.5°C & higher intensity of heavy precipitation events & in most regions in Europe & medium confidence \\ \hline
\end{tabular}
\caption{Example of spatiotemporal contextualization - alignment with confidence label.}
\label{app:sample-8}
\end{table}

\subsubsection{No Quantifiers \textit{(--No\_Quantifier)}} \label{no-quantifiers}

\paragraph{General rule} You should be able to read the causal relation between [\textit{Cause--No\_Quantifier}] and [\textit{Effect--No\_Quantifier}] as either:
\begin{itemize}
    \item \textit{``An increase in} [\textit{Cause--No\_Quantifier}] \textit{causes an \underline{increase} in} [\textit{Effect--No\_Quantifier}]\textit{''};
    \item or, \textit{``An increase in} [\textit{Cause--No\_Quantifier}] \textit{causes a \underline{decrease} in} [\textit{Effect--No\_Quantifier}]\textit{''}.
\end{itemize}

\paragraph{Reformulation} Make minor alterations to the original phrasing of the cause and effect in [\textit{Cause--NP}] and [\textit{Effect--NP}], which are then included as reformulations in [\textit{Cause--No\_Quantifier}] and [\textit{Effect--No\_Quantifier}]. 

\begin{itemize}
    \item Remove adjectives, adverbs, and nouns that signal the \textbf{direction} of the effect:
    \begin{itemize}
        \item \underline{Increasingly} irreversible losses in cryospheric ecosystems [\textit{Effect--NP}]; irreversible losses in cryospheric ecosystems [\textit{Effect--No\_Quantifier}].
        \item \underline{Reduced} food security [\textit{Effect--NP}]; food security [\textit{Effect--No\_Quantifier}].
        \item \underline{Reduction of} cryospheric elements [\textit{Effect--NP}]; cryospheric elements [\textit{Effect--No\_Quantifier}].
    \end{itemize}
    \item Remove adjectives, adverbs, and nouns that signal an \textbf{aspect} of the effect:
    \begin{itemize}
        \item \underline{Affected} water security [\textit{Effect--NP}]; water security [\textit{Effect--No\_Quantifier}].
        \item \underline{Hindered} efforts to meet Sustainable Development Goals [\textit{Effect--NP}]; efforts to meet Sustainable Development Goals [\textit{Effect--No\_Quantifier}].
        \item \underline{Severe} water scarcity [\textit{Effect--NP}]; water scarcity [\textit{Effect--No\_Quantifier}].
    \end{itemize}
\end{itemize}

\textit{``Anthropogenic''} is reformulated as \textit{``human activities''} [\textit{Cause--NP}].

\subsection{Explicitness} \label{explicitness}
Write ``E'' or ``I'' in the [\textit{Explicitness}] column. The distinction between explicit and implicit is as follows:
\begin{itemize}
    \item Implicit (causal relation is not obvious from the target):
    \begin{itemize}
        \item The [\textit{Target}] is ``/''.
        \item The [\textit{Target}] is not a ``standard'' causation-invoking verb like \textit{``cause''}, \textit{``affect''}, and \textit{``increase''}. Examples are \textit{``expose''}, \textit{``cope with''}, \textit{``encourage''}.
        \item Causality is embedded in a word, abbreviation, or adjective (e.g., AFOLU, anthropogenic).
    \end{itemize}
    \item Explicit otherwise.
\end{itemize}

\subsection{Correlation} \label{correlation}
[\textit{Correlation}] reflects the direction of the correlation between [\textit{Cause--No\_Quantifier}] and [\textit{Effect--No\_Quantifier}].
\begin{itemize}
    \item The [\textit{Correlation}] = \textit{Positive}, when:
    \begin{itemize}
        \item \textit{``An increase in} [\textit{Cause--No\_Quantifier}] \textit{causes an increase in} [\textit{Effect--No\_Quantifier}]\textit{''};
        \item \textit{``A decrease in} [\textit{Cause--No\_Quantifier}] \textit{causes a decrease in} [\textit{Effect--No\_Quantifier}]\textit{''}.
    \end{itemize}
    \item The [\textit{Correlation}] = \textit{Negative}, when:
    \begin{itemize}
        \item \textit{``An increase in} [\textit{Cause--No\_Quantifier}] \textit{causes a decrease in} [\textit{Effect--No\_Quantifier}]\textit{''};
        \item \textit{``A decrease in} [\textit{Cause--No\_Quantifier}] \textit{causes an increase in} [\textit{Effect--No\_Quantifier}]\textit{''}.
    \end{itemize}
\end{itemize}

\subsection{Relation Type} \label{relation type}

[\textit{Relation type}] reflects the type of relation between [\textit{Cause--No\_Quantifier}] and [\textit{Effect--No\_Quantifier}].
\begin{itemize}
    \item The [\textit{Relation type}] = \textit{Positive}, when \textit{``The existence of} [\textit{Cause--No\_Quantifier}] \textit{\underline{leads to} the existence of} [\textit{Effect--No\_Quantifier}]\textit{''}.
    \item The [\textit{Correlation}] = \textit{Negative}, when \textit{``The existence of} [\textit{Cause--No\_Quantifier}] \textit{\underline{prevents} the existence of} [\textit{Effect--No\_Quantifier}]\textit{''}.
\end{itemize}

A popular target that signals a negative relation type is \textit{``avoid''}.

\subsection{Abbreviations} \label{abbreviations}

If the statement mentions an abbreviation, e.g., \textit{CO\textsubscript{2}}, then this should be resolved in all columns except [\textit{Target}] and included as a set in [\textit{Abbreviations}]. [\textit{Abbreviations}] should be structured as follows, for example, for GHG: [\textit{Abbreviations}] = \textit{GHG = greenhouse gases}. In case two or more abbreviations are used (e.g., GHG and AFOLU), then [\textit{Abbreviations}] = \{\textit{GHG = greenhouse gases; AFOLU = Land Use, Land Use Change, and Forestry}\} (mention them in the order of appearance in the statement).

\paragraph{Note for [\textit{Target}]} Abbreviations should be resolved in all columns EXCEPT [\textit{Target}]; in that column you can use the abbreviation as is.

\section{Overview Resolved Abbreviations} \label{appendix:abbreviations}

Table \ref{tab:abbreviations} shows an overview of all resolved abbreviations.

\begin{table}[ht!]
    \centering
    \small
    \begin{tabular}{p{1.5cm}p{5.5cm}}
    \toprule
        Abbr. & Meaning \\
    \midrule
        AFOLU & Agriculture, Forestry, and Other Land Use \\
        CH4 & Methane \\
        CO2 & Carbon dioxode \\
        CO2-FFI & Carbon dioxide from fossil fuels and industrial processes \\
        CO2-LULUCF & Carbon dioxide emissions from land use, land-use change and forestry \\
        GHG & Greenhouse gases \\
        GDP & Gross domestic product \\
        LDC & Least Developed Countries \\
        N2O & Nitrous oxide \\
        O3 & Tropospheric ozone \\
        SIDS & Small Island Developing States \\
        SLCF & Short-Lived Climate Forcers \\
    \bottomrule
    \end{tabular}
    \caption{Twelve abbreviations used in IPCC that occur in \textit{ClimateCause}.}
    \label{tab:abbreviations}
\end{table}

\section{Annotation Description and Disagreement Resolution} \label{appendix:annotation_round_1}

\subsection{Annotator Description}

Both annotators are within the [25-30 years] age range, female, non-native speakers of English (C1 proficiency), and highly educated. They had been involved in another causality annotation campaign prior to this study. Annotator A is European, has a PhD, and holds a master's degree in linguistics (English). Annotator B is South American, pursues a PhD in the field of linguistics and communication, and has a master's degree in the humanities. The third annotator consulted during disagreement resolution is within the [30-40 years] age range, non-binary, non-native speaker of English (C1/C2 proficiency), and highly educated (PhD). No one has an academic background in environmental science.

\subsection{Disagreement Resolution}
We followed the following procedure to resolve disagreement in the annotations during the first annotation round:
\begin{enumerate}
    \item Manually identify the causal relations that both annotators retrieved.
    \item Compare annotations for these relations:
    \begin{enumerate}
        \item Mark and resolve incorrect spans.
        \item Mark and correct violations against annotation guidelines.
        \item Mark disagreement between annotators.
    \end{enumerate}
    \item Look at the causal relations that one annotator annotated but the second did not:
    \begin{enumerate}
        \item Include relations that the second annotator did not include but that are very similar to other relation they annotated before, e.g., anthropogenic. They possibly missed these due to a high level of cognitive load of the task.
        \item Include relations that the missed annotator did not include, but indicated in the comments that they were not sure about this relation.
        \item Include relations that are implied in abbreviations. The lack of annotating the relations is presumed to be caused by a lack of knowledge. The abbreviations are checked. 
        \item Include causal relations that are not annotated due to violations to the annotation guidelines, e.g., \textit{reductions in CH4 and other ozone precursors} not split in two events.
    \end{enumerate}
    \item For remaining unresolved relations:
    \begin{enumerate}
        \item Consult third annotator, then majority vote.
    \end{enumerate}
\end{enumerate}

\section{Readability of IPCC Reports} \label{appendix:readability_metrics}

IPCC reports are known for low readability \citep{barkemeyer2016linguistic}. We examine whether statements in the ClimateCause dataset are low in readability and whether they are less readable than those in related climate causality datasets \citep{pineda2025polaris3, pineda2025polaris4}. 
We measure readability using five established metrics: Flesch Reading Ease (FRE) \citep{flesch1948new}, Flesch-Kincaid Grade Level (FKG) \citep{kincaid1975derivation}, Automated Readability Index (ARI) \citep{kincaid1975derivation}, Coleman-Liau Index (CLI) \citep{coleman1975computer}, and Dale-Chall Readability Score (DCRS) \citep{chall1995readability}. Formulae are in Table \ref{tab:readability_formula}. 

\begin{figure}[t!]
\centering
\begin{tikzpicture}
\begin{axis}[
    ybar,
    bar width=6pt,
    width=8cm,
    height=3cm,
    enlarge x limits=0.2,
    ylabel={Mean Value},
    xlabel style={font=\scriptsize}, 
    ylabel style={font=\scriptsize},
    symbolic x coords={FRE ($\uparrow$), FKG ($\downarrow$), CLI ($\downarrow$), ARI ($\downarrow$), DCRS ($\downarrow$)},
    xtick=data,
    axis x line=bottom,
    axis y line=left,
    ymin=0,
    ymax=60, 
    legend style={at={(0.5,-0.35)}, anchor=north, legend columns=-1, font=\scriptsize},
    nodes near coords,
    every node near coord/.append style={font=\tiny, rotate=90, anchor=west},
    tick label style={font=\scriptsize},
    enlarge x limits=0.15
]
\addplot[fill=gray!10] coordinates {(FRE ($\uparrow$),52.25) (FKG ($\downarrow$),10.48) (CLI ($\downarrow$),10.91) (ARI ($\downarrow$),8.93) (DCRS ($\downarrow$),10.56)};
\addplot[fill=gray!90] coordinates {(FRE ($\uparrow$),49.71) (FKG ($\downarrow$),10.56) (CLI ($\downarrow$),15.69) (ARI ($\downarrow$),13.97) (DCRS ($\downarrow$),11.16)};
\addplot[fill=blue] coordinates {(FRE ($\uparrow$),20.72) (FKG ($\downarrow$),16.20) (CLI ($\downarrow$),16.03) (ARI ($\downarrow$),18.36) (DCRS ($\downarrow$),13.17)};
\legend{PolarIs3CAUS, PolarIs4CAUS, \textit{ClimateCause}}
\end{axis}
\end{tikzpicture}
\caption{Mean readability scores of statements in PolarIs3CAUS, and PolarIs4CAUS, and \textit{ClimateCause} (ours). Arrows indicate direction of higher readability.}
\label{fig:readability-bar}
\end{figure}
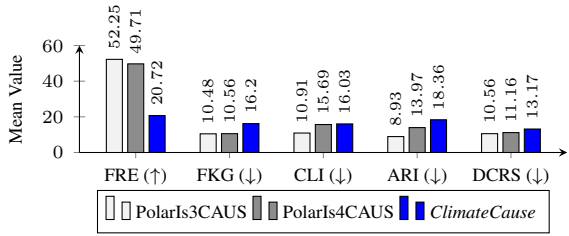

Results confirm low readability in \textit{ClimateCause}. Most statements require college-level reading, with half rated very difficult (FRE $\in [10,30]$) or extremely difficult (FRE $\in [0,10]$). They are also less readable than those in climate causality datasets from social media \citep{pineda2025polaris3, pineda2025polaris4} (Figure \ref{fig:readability-bar}; Table \ref{tab:readability}). 

\begin{table*}[ht]
    \centering
    \small
    \begin{tabular}{l|l}
    \toprule
        \textbf{Metric} & \textbf{Formula} \\
    \midrule
        Flesch Reading Ease & $206.835 - 1.015 \left( \frac{\text{total words}}{\text{total sentences}} \right) - 84.6 \left( \frac{\text{total syllables}}{\text{total words}} \right)$ \\
        & \\
        Flesch-Kincaid Grade Level & $0.39 \left( \frac{\text{total words}}{\text{total sentences}} \right) + 11.8 \left( \frac{\text{total syllables}}{\text{total words}} \right) - 15.59$ \\
& \\
        Automated Readability Index & $4.71 \left( \frac{\text{characters}}{\text{words}} \right) + 0.5 \left( \frac{\text{words}}{\text{sentences}} \right) - 21.43$ \\
& \\
        Coleman-Liau Index & $0.0588 \left( \frac{\text{letters}}{\text{words}} \right) \times 100
- 0.296 \left( \frac{\text{sentences}}{\text{words}} \right) \times 100 - 15.8$ \\
& \\
        Dale-Chall Readability & $0.1579 \left( \frac{\text{difficult words}}{\text{words}} \times 100 \right) + 0.0496 \left( \frac{\text{words}}{\text{sentences}} \right)$ \\
    \bottomrule
    \end{tabular}
    \caption{Readability metrics and their formula.}
    \label{tab:readability_formula}
\end{table*}

\begin{table}[t!]
\centering
\footnotesize
\begin{tabular}{|p{1.1cm}|p{0.8cm}p{0.8cm}p{0.8cm}p{0.9cm}p{0.8cm}|}
\hline
\textbf{Metric} & \textbf{Median} & \textbf{Mean} & \textbf{Std} & \textbf{Min} & \textbf{Max} \\
\hline
\multicolumn{6}{|c|}{\textbf{ClimateCause} (Science-for-policy reports)} \\
\hline
FRE ($\uparrow$) & $\mathbf{26.95}$ & $\mathbf{20.72}$ & $\mathbf{30.39}$ & $\mathbf{-67.15}$ & $\mathbf{93.26}$ \\
FKG ($\downarrow$) & $\mathbf{16.06}$ & $\mathbf{16.20}$ & $\mathbf{6.05}$ & $\mathbf{1.80}$ & $\mathbf{30.27}$ \\
CLI ($\downarrow$) & $\mathbf{15.97}$ & $\mathbf{16.03}$ & $\mathbf{5.01}$ & $\mathbf{1.71}$ & $\mathbf{25.72}$ \\
ARI ($\downarrow$) & $\mathbf{18.76}$ & $\mathbf{18.36}$ & $\mathbf{6.97}$ & $\mathbf{2.20}$ & $\mathbf{34.84}$ \\
DCRS ($\downarrow$) & $\mathbf{13.21}$ & $\mathbf{13.17}$ & $\mathbf{1.34}$ & $\mathbf{9.92}$ & $\mathbf{15.62}$ \\
\hline
\multicolumn{6}{|c|}{\textbf{PolarIs3CAUS} (Reddit)} \\
\hline
FRE ($\uparrow$) & $53.64$ & $52.25$ & $21.66$ & $-1.28$ & $103.54$ \\
FKG ($\downarrow$) & $10.45$ & $10.48$ & $4.28$ & $0.80$ & $20.55$ \\
CLI ($\downarrow$) & $10.36$ & $10.91$ & $4.35$ & $1.53$ & $23.17$ \\
ARI ($\downarrow$) & $8.77$ & $8.93$ & $5.21$ & $-3.30$ & $21.74$ \\
DCRS ($\downarrow$) & $10.43$ & $10.56$ & $2.01$ & $6.01$ & $16.52$ \\
\hline
\multicolumn{6}{|c|}{\textbf{PolarIs4CAUS} (Twitter/X)} \\
\hline
FRE ($\uparrow$) & $47.79$ & $49.71$ & $23.70$ & $-9.48$ & $103.54$ \\
FKG ($\downarrow$) & $10.32$ & $10.56$ & $4.82$ & $0.52$ & $26.10$ \\
CLI ($\downarrow$) & $15.20$ & $15.69$ & $5.39$ & $2.98$ & $31.34$ \\
ARI ($\downarrow$) & $13.33$ & $13.97$ & $5.73$ & $1.41$ & $34.89$ \\
DCRS ($\downarrow$) & $11.47$ & $11.16$ & $2.07$ & $5.50$ & $15.26$ \\
\hline
\end{tabular}
\caption{Readability scores of statements in ClimateCause (ours) from science-for-policy reports, PolarIs3CAUS \citep{pineda2025polaris3} from Reddit discussions, and PolarIs4CAUS \citep{pineda2025polaris4} from Twitter/X posts. Arrows indicate direction for higher readability ($\uparrow$: higher score means higher readability; $\downarrow$: lower score means higher readability).}
\label{tab:readability}
\end{table}

\section{Readability of Reported Causality: Examples} \label{appendix:readability_metrics_reported_causality}

\subsection{Co-occurrence between Metrics}

The coincidence matrix illustrates the co-occurrence between complexity metrics in \textit{ClimateCause}:
\begin{figure}[ht!]
    \centering
        \footnotesize
        \begin{tabular}{p{0.6cm}|p{0.3cm}p{0.3cm}p{0.3cm}p{0.3cm}p{0.3cm}}
            ${com}$ & \textit{10} & &  &  &  \\
            ${ex}$ & 2 & \textit{17} &  &  &  \\
            ${nest}$ & 2 & 5 & \textit{12} &  &  \\ 
            ${corr}$ & 4 & 8 & 4 & \textit{24} &  \\
            ${pol}$ & 1 & 2 & - & 3 & \textit{3} \\
            \midrule
            & ${com}$ & ${ex}$ & ${nest}$ & ${corr}$ & ${pol}$ \\
        \end{tabular}
    \label{fig:combined}
\end{figure}

\subsection{Statements with High Complexity Scores} \label{appendix:high_complexity}

\paragraph{Overarching cause/effect} $C^{com}(s) = 12$. Statement: \textit{``Similarly, integrated transport and energy infrastructure planning and operations can together reduce the environmental, social, and economic impacts of decarbonising the transport and energy sectors.''}

\paragraph{Examples} $C^{ex}(s) = 16$. Statement: \textit{``Accelerated support from developed countries and multilateral institutions is a critical enabler to enhance mitigation and adaptation action and can address inequities in finance, including its costs, terms and conditions, and economic vulnerability to climate change. Scaled-up public grants for mitigation and adaptation funding for vulnerable regions, e.g., in Sub-Saharan Africa, would be cost-effective and have high social returns in terms of access to basic energy. Options for scaling up mitigation and adaptation in developing regions include: increased levels of public finance and publicly mobilised private finance flows from developed to developing countries in the context of the USD 100 billion-a-year goal of the Paris Agreement; increase the use of public guarantees to reduce risks and leverage private flows at lower cost; local capital markets development; and building greater trust in international cooperation processes. A coordinated effort to make the post-pandemic recovery sustainable over the long term through increased flows of financing over this decade can accelerate climate action, including in developing regions facing high debt costs, debt distress and macroeconomic uncertainty.''}

\paragraph{Nested causality} $C^{nest}(s) = 20.93$. Statement: \textit{``Human-caused climate change is a consequence of more than a century of net GHG emissions from energy use, land-use and land use change, lifestyle and patterns of consumption, and production. Emissions reductions in CO2 from fossil fuels and industrial processes (CO2-FFI), due to improvements in energy intensity of GDP and carbon intensity of energy, have been less than emissions increases from rising global activity levels in industry, energy supply, transport, agriculture and buildings. The 10\% of households with the highest per capita emissions contribute 34–45\% of global consumption-based household GHG emissions, while the middle 40\% contribute 40–53\%, and the bottom 50\% contribute 13–15\%. An increasing share of emissions can be attributed to urban areas (a rise from about 62\% to 67–72\% of the global share between 2015 and 2020). The drivers of urban GHG emissions are complex and include population size, income, state of urbanisation and urban form.}

\paragraph{Correlation} $C^{corr}(s) = 70$. Statement: \textit{``Human-caused climate change is already affecting many weather and climate extremes in every region across the globe. This has led to widespread adverse impacts on food and water security, human health and on economies and society and related losses and damages to nature and people.''}

\paragraph{Relation type} $C^{pol}(s) = 40$. Statement: \textit{``Trade-offs in terms of employment, water use, land-use competition and biodiversity, as well as access to, and the affordability of, energy, food, and water can be avoided by well-implemented land-based mitigation options, especially those that do not threaten existing sustainable land uses and land rights, with frameworks for integrated policy implementation.''}

\section{Benchmarking Causal Reasoning} \label{appendix:benchmarking}

\subsection{Data and Code Availability}

\paragraph{Data} We released \textit{ClimateCause} as a CSV-formatted file in a dedicated GitHub repository under the CC-BY 4.0 license, which is intended for use in academic and research settings: \url{https://github.com/laallein/ClimateCause}.

\paragraph{Code for data retrieval} We released the Python scripts for retrieving the IPCC statements from the Wikibase (SQL).

\paragraph{Code for readability/benchmarking} We released the Python scripts for running the readability analyses with the proposed complexity metrics regarding the readability of reported causality in text and the benchmarking experiments for correlation inference and causal chain reasoning under an open-source license in the GitHub repository. 

\subsection{Implementation Details}

We evaluate causal reasoning abilities in GPT5.1; version \texttt{gpt-5.1-2025-11-13}, which we access through the OpenAI API. We therefore rely on the OpenAI hardware facilities for running the experiments. We use the default hyperparameter settings. The model's context window size is 400,000; its maximum number of output tokens is 128,000, and its knowledge cutoff date is 30 September 2024. The experiments were sent to the Batch API, where in total 34,303 requests were made, such that the model handled 9,880,620 completion tokens (Costs: \$5.052 input tokens; \$11.4 output tokens). 

\subsection{Evaluation Metrics}

\paragraph{Precision}
\begin{equation}
    \text{Precision} = \frac{TP}{TP+FP}
\end{equation}

\paragraph{Recall}
\begin{equation}
    \text{Recall} = \frac{TP}{TP+FN}
\end{equation}

\paragraph{F1-score}
\begin{equation}
    \text{F1} = \frac{2 \times \text{precision} \times \text{recall}}{\text{precision} + \text{recall}}
\end{equation}
\subsection{Breakdown of Results}

Per-class performance for chain position identification is reported in Table \ref{tab:per-class-performance}. A full breakdown of performance with each individual prompt is given in Table \ref{tab:corrI-results} for correlation inference and in Table \ref{tab:ccr-results} for causal chain reasoning. 

\begin{table}[ht]
    \centering
    \footnotesize
    \begin{tabular}{p{0.8cm}|p{1.6cm}p{1.6cm}p{1.6cm}}
    \toprule
    \textbf{Class} & \textbf{Precision} & \textbf{Recall} & \textbf{F1} \\
\midrule
\multicolumn{4}{c}{\textbf{CCR \textit{position}}} \\
\midrule
none   & $0.9915_{\pm 0.0082}$ & $0.5357_{\pm 0.1814}$ & $0.6800_{\pm 0.1496}$ \\
start  & $0.2687_{\pm 0.0551}$ & $0.8438_{\pm 0.0950}$ & $0.4055_{\pm 0.0680}$ \\
middle & $0.5008_{\pm 0.1714}$ & $0.8565_{\pm 0.0887}$ & $0.6126_{\pm 0.1191}$ \\
end    & $0.3123_{\pm 0.1877}$ & $0.7619_{\pm 0.1400}$ & $0.4291_{\pm 0.2027}$ \\
\midrule
\multicolumn{4}{c}{\textbf{CCR+ECI+RC \textit{position}}} \\
\midrule
none   & $0.8842_{\pm 0.0167}$ & $0.5679_{\pm 0.1402}$ & $0.6818_{\pm 0.1065}$ \\
start  & $0.1934_{\pm 0.0632}$ & $0.4931_{\pm 0.1752}$ & $0.2665_{\pm 0.0818}$ \\
middle & $0.2788_{\pm 0.0631}$ & $0.4769_{\pm 0.1726}$ & $0.3317_{\pm 0.0604}$ \\
end    & $0.2283_{\pm 0.0690}$ & $0.4762_{\pm 0.1340}$ & $0.3001_{\pm 0.0813}$ \\
\bottomrule
    \end{tabular}
    \caption{Per-class performance for chain position identification.}
    \label{tab:per-class-performance}
\end{table}

\begin{table}[ht]
\centering
\footnotesize
\begin{tabular}{p{2.7cm}|p{1cm}p{1cm}p{1cm}}
\toprule
\textbf{Prompting Strategy} & \textbf{Precision} & \textbf{Recall} & \textbf{F1} \\
\midrule
\multicolumn{4}{c}{\textbf{CorrI}} \\
\midrule
CorrI\_0\_1       & $0.7877$ & $0.9449$ & $0.8592$ \\
CorrI\_0\_2       & $0.7197$ & $0.8881$ & $0.7951$ \\
CorrI\_0\_3       & $0.9538$ & $0.5680$ & $0.7120$ \\
CorrI\_F\_1       & $0.9284$ & $0.9593$ & $0.9436$ \\
CorrI\_F\_2       & $0.7716$ & $0.9243$ & $0.8410$ \\
CorrI\_F\_3       & $0.9621$ & $0.9621$ & $0.9621$ \\
CorrI\_CoT\_1     & $0.8565$ & $0.9552$ & $0.9032$ \\
CorrI\_CoT\_2     & $0.7580$ & $0.8950$ & $0.8208$ \\
CorrI\_CoT\_3     & $0.9522$ & $0.9604$ & $0.9563$ \\
\midrule
\multicolumn{4}{c}{\textbf{CorrI+RC}} \\
\midrule
CorrI\_RC\_0\_1   & $0.7491$ & $0.7091$ & $0.7286$ \\
CorrI\_RC\_0\_2   & $0.7561$ & $0.8537$ & $0.8019$ \\
CorrI\_RC\_0\_3   & $0.9830$ & $0.5955$ & $0.7417$ \\
CorrI\_RC\_F\_1   & $0.9803$ & $0.9681$ & $0.9742$ \\
CorrI\_RC\_F\_2   & $0.8683$ & $0.9415$ & $0.9034$ \\
CorrI\_RC\_F\_3   & $0.9794$ & $0.9828$ & $0.9811$ \\
CorrI\_RC\_CoT\_1 & $0.8986$ & $0.7935$ & $0.8428$ \\
CorrI\_RC\_CoT\_2 & $0.7641$ & $0.8864$ & $0.8207$ \\
CorrI\_RC\_CoT\_3 & $0.9236$ & $0.9570$ & $0.9400$ \\
\bottomrule
\end{tabular}
\caption{Performance results for CorrI strategies.}
\label{tab:corrI-results}
\end{table}

\begin{table}[ht]
\centering
\footnotesize
\begin{tabular}{p{3.1cm}|p{1cm}p{1cm}p{1cm}}
\toprule
\textbf{Prompting Strategy} & \textbf{Precision} & \textbf{Recall} & \textbf{F1} \\
\midrule
\multicolumn{4}{c}{\textbf{CCR \textit{member}}} \\
\midrule
CCR\_member\_A\_4   & $0.9083$ & $0.8609$ & $0.8839$ \\
CCR\_member\_A\_5   & $0.3551$ & $0.8522$ & $0.5013$ \\
CCR\_member\_A\_6   & $0.3512$ & $0.9130$ & $0.5072$ \\
CCR\_member\_SN\_4  & $0.8043$ & $0.9652$ & $0.8775$ \\
CCR\_member\_SN\_5  & $0.3631$ & $0.9913$ & $0.5315$ \\
CCR\_member\_SN\_6  & $0.3394$ & $0.9652$ & $0.5023$ \\
CCR\_member\_ML\_4  & $1.0000$ & $0.8174$ & $0.8995$ \\
CCR\_member\_ML\_5  & $0.5067$ & $0.9913$ & $0.6706$ \\
CCR\_member\_ML\_6  & $0.4197$ & $1.0000$ & $0.5913$ \\
\midrule
\multicolumn{4}{c}{\textbf{CCR \textit{position}}} \\
\midrule
CCR\_position\_A\_4  & $0.3784$ & $0.9655$ & $0.5437$ \\
CCR\_position\_A\_5  & $0.2348$ & $0.9643$ & $0.3776$ \\
CCR\_position\_A\_6  & $0.2651$ & $0.8462$ & $0.4037$ \\
CCR\_position\_SN\_4 & $0.3919$ & $1.0000$ & $0.5631$ \\
CCR\_position\_SN\_5 & $0.2295$ & $1.0000$ & $0.3733$ \\
CCR\_position\_SN\_6 & $0.2527$ & $0.8846$ & $0.3932$ \\
CCR\_position\_ML\_4 & $0.3404$ & $1.0000$ & $0.5079$ \\
CCR\_position\_ML\_5 & $0.2203$ & $1.0000$ & $0.3611$ \\
CCR\_position\_ML\_6 & $0.2393$ & $0.9655$ & $0.3836$ \\
\midrule
\multicolumn{4}{c}{\textbf{CCR+ECI+RC \textit{member}}} \\
\midrule
CCR\_ECI\_member\_0\_4    & $0.2774$ & $0.9478$ & $0.4291$ \\
CCR\_ECI\_member\_0\_5    & $0.2873$ & $0.8870$ & $0.4340$ \\
CCR\_ECI\_member\_0\_6    & $0.2933$ & $0.8696$ & $0.4386$ \\
CCR\_ECI\_member\_F\_4    & $0.3363$ & $0.6609$ & $0.4457$ \\
CCR\_ECI\_member\_F\_5    & $0.2620$ & $0.9043$ & $0.4063$ \\
CCR\_ECI\_member\_F\_6    & $0.5000$ & $0.5043$ & $0.5022$ \\
CCR\_ECI\_member\_CoT\_4  & $0.4839$ & $0.6522$ & $0.5556$ \\
CCR\_ECI\_member\_CoT\_5  & $0.2875$ & $0.8000$ & $0.4230$ \\
CCR\_ECI\_member\_CoT\_6  & $0.2881$ & $0.9043$ & $0.4370$ \\
\midrule
\multicolumn{4}{c}{\textbf{CCR+ECI+RC \textit{position}}} \\
\midrule
CCR\_ECI\_position\_0\_4   & $0.1942$ & $0.7692$ & $0.3101$ \\
CCR\_ECI\_position\_0\_5   & $0.1944$ & $0.7241$ & $0.3066$ \\
CCR\_ECI\_position\_0\_6   & $0.1667$ & $0.5185$ & $0.2523$ \\
CCR\_ECI\_position\_F\_4   & $0.1429$ & $0.6774$ & $0.2360$ \\
CCR\_ECI\_position\_F\_5   & $0.1585$ & $0.5417$ & $0.2453$ \\
CCR\_ECI\_position\_F\_6   & $0.2927$ & $0.4444$ & $0.3529$ \\
CCR\_ECI\_position\_CoT\_4 & $0.3167$ & $0.6333$ & $0.4222$ \\
CCR\_ECI\_position\_CoT\_5 & $0.2727$ & $0.6429$ & $0.3830$ \\
CCR\_ECI\_position\_CoT\_6 & $0.1905$ & $0.2500$ & $0.2162$ \\
\bottomrule
\end{tabular}
\caption{Performance results for CCR strategies.}
\label{tab:ccr-results}
\end{table}

\subsection{Prompts}

We provide all prompts for correlation inference in Table \ref{tab:prompts_corrI} (CorrI) and \ref{tab:prompts_corrI_RC} (CorrI+RC), and for causal chain reasoning in Table \ref{tab:prompts_ccr} (CCR, membership) \ref{tab:prompts_ccr_position} (CCR, position), \ref{tab:prompts_ccr_eci} (CCR+ECI+RC, membership), \ref{tab:prompts_ccr_eci_position_part1} and \ref{tab:prompts_ccr_eci_position_part2} (CCR+ECI+RC, position). 

\subsubsection{Examplar Selection for Few-Shot Setting} The following examples are taken from the IPCC report from which the statements in \textit{ClimateCause} are drawn. We made sure to select them from different sections and/or paragraphs and with causal relations that are not included in the dataset.

\paragraph{Example 1} \underline{Statement}: \textit{``Climate resilient development is enabled by increased international cooperation including mobilising and enhancing access to finance, particularly for developing countries, vulnerable regions, sectors and groups and aligning finance flows for climate action to be consistent with ambition levels and funding needs.''} \underline{Causal relation}: \textit{access to finance} $\rightarrow$ \textit{climate resilient development}. \underline{Correlation}: \textit{positive} [CorrI-F, CorrI+RC-F]. \underline{Motivation}: Switched order of cause and effect in text; cause and effect are not next to each other; and cause is an example of overarching event in the statement (i.e., increased international cooperation). \underline{Position in IPCC report}: 4.8.2. International Cooperation and Coordination, p 112.

\paragraph{Example 2} \underline{Statement}: \textit{``Persistent and region-specific barriers also continue to hamper the economic and political feasibility of deploying AFOLU mitigation options.''} \underline{Causal relation}: \textit{persistent barriers} $\rightarrow$ \textit{political feasibility of deploying AFOLU mitigation options}. \underline{Correlation}: \textit{negative} [CorrI-F, CorrI+RC-F]. \underline{List of events}: [\textit{persistent barriers}, \textit{region-specific barriers}, \textit{economic feasibility of deploying AFOLU mitigation options}, \textit{political feasibility of deploying AFOLU mitigation options}] [CCR\_ECI\_member\_F, CCR\_ECI\_position\_F]. \underline{Motivation}: Split of NPs; mention of AFOLU; long effect formulation; no causal chains. \underline{Position in IPCC report}: 2.3.1. The Gap Between Mitigation Policies, Pledges and Pathways that Limit Warming to 1.5°C or Below 2°C, p 61.

\paragraph{Example 3} \underline{Statement}: \textit{``Adaptation options can become maladaptive due to their environmental impacts that constrain ecosystem services and decrease biodiversity and ecosystem resilience to climate change or by causing adverse outcomes for different groups, exacerbating inequity.''} \underline{List of events}: [\textit{adaptation options}, \textit{environmental impacts}, \textit{ecosystem services}, \textit{biodiversity}, \textit{ecosystem resilience to climate change}, \textit{adverse outcomes for different groups}, \textit{inequity}, \textit{maladaptive adaptation options}] [CCR\_ECI\_member\_F, CCR\_ECI\_position\_F]. \underline{Motivation}: Low readability statement, many causal relations, two chains, start and end of chain not in their relative position in the statement (i.e., end of chain is mentioned before start). \underline{Position in IPCC report}: 3.2 Long-term Adaptation Options and Limits, p 78.

\section{Statistical Testing}

The results of the McNemar tests between CorrI and CorrI+RC are in Table \ref{mcnemar-corrI}, between CCR member and position in Table \ref{mcnemar-CCR}, and between CCR+ECI+RC member and position in Table \ref{mcnemar-CCR+ECI+RC}.

\begin{table}[t]
    \centering
    \footnotesize
    \begin{subtable}[t]{0.48\textwidth}
        \centering
        \begin{tabular}{p{1.0cm}p{1.0cm}p{1.3cm}p{2cm}p{0.3cm}}
            \toprule
            \textbf{CorrI} & \textbf{CorrI+RC} & \textbf{$\chi^2$} & \textbf{p-value} & \\
            \midrule
            0\_1 & 0\_1 & 75.3216 & $3.9997 \times 10^{-18}$ & * \\
            0\_2 & 0\_2 & 13.8996 & $1.9284 \times 10^{-4}$ & * \\
            0\_3 & 0\_3 & 0.1147 & $7.3488 \times 10^{-1}$ & \\
            F\_1 & F\_1 & 8.0152 & $4.6388 \times 10^{-3}$ & * \\
            F\_2 & F\_2 & 20.3125 & $6.5770 \times 10^{-6}$ & * \\
            F\_3 & F\_3 & 0.0000 & $1.0000$ & \\
            CoT\_1 & CoT\_1 & 84.3005 & $4.2501 \times 10^{-20}$ & * \\
            CoT\_2 & CoT\_2 & 0.5042 & $4.7768 \times 10^{-1}$ & \\
            CoT\_3 & CoT\_3 & 2.9605 & $8.5320 \times 10^{-2}$ & \\
            \bottomrule
        \end{tabular}
        \caption{CorrI vs CorrI+RC.}
        \label{mcnemar-corrI}
    \end{subtable}
    \hfill
    \begin{subtable}[t]{0.48\textwidth}
        \centering
        \begin{tabular}{p{1.0cm}p{1.0cm}p{1.3cm}p{2cm}p{0.3cm}}
            \toprule
            \textbf{CCR member} & \textbf{CCR position} & \textbf{$\chi^2$} & \textbf{p-value} & \\
            \midrule
            A\_4 & A\_4 & 41.4902 & $1.1846 \times 10^{-10}$ & * \\
            A\_5 & A\_5 & 27.8409 & $1.3171 \times 10^{-7}$ & * \\
            A\_6 & A\_6 & 11.5056 & $6.9386 \times 10^{-4}$ & * \\
            SN\_4 & SN\_4 & 20.0000 & $7.7442 \times 10^{-6}$ & * \\
            SN\_5 & SN\_5 & 13.7931 & $2.0408 \times 10^{-4}$ & * \\
            SN\_6 & SN\_6 & 19.5556 & $9.7716 \times 10^{-6}$ & * \\
            ML\_4 & ML\_4 & 65.1268 & $7.0232 \times 10^{-16}$ & * \\
            ML\_5 & ML\_5 & 45.1765 & $1.8006 \times 10^{-11}$ & * \\
            ML\_6 & ML\_6 & 0.1023 & $7.4912 \times 10^{-1}$ & \\
            \bottomrule
        \end{tabular}
        \caption{CCR membership vs CCR position.}
        \label{mcnemar-CCR}
    \end{subtable}
    \hfill
    \begin{subtable}[t]{0.48\textwidth}
        \centering
        \begin{tabular}{p{1.2cm}p{1.2cm}p{1cm}p{2cm}p{0.2cm}}
            \toprule
            \textbf{CCR+ECI +RC member} & \textbf{CCR+ECI +RC position} & \textbf{$\chi^2$} & \textbf{p-value} & \\
            \midrule
            0\_4 & 0\_4 & 99.3103 & $2.1588 \times 10^{-23}$ & * \\
            0\_5 & 0\_5 & 59.5044 & $1.2202 \times 10^{-14}$ & * \\
            0\_6 & 0\_6 & 101.1500 & $8.5275 \times 10^{-24}$ & * \\
            F\_4 & F\_4 & 27.0096 & $2.0245 \times 10^{-7}$ & * \\
            F\_5 & F\_5 & 37.8125 & $7.7881 \times 10^{-10}$ & * \\
            F\_6 & F\_6 & 0.1552 & $6.9364 \times 10^{-1}$ & \\
            CoT\_4 & CoT\_4 & 1.1228 & $2.8931 \times 10^{-1}$ & \\
            CoT\_5 & CoT\_5 & 104.6639 & $1.4471 \times 10^{-24}$ & * \\
            CoT\_6 & CoT\_6 & 129.0076 & $6.7558 \times 10^{-30}$ & * \\
            \bottomrule
        \end{tabular}
        \caption{CCR+ECI+RC membership vs position.}
        \label{mcnemar-CCR+ECI+RC}
    \end{subtable}
    \caption{McNemar tests across different label set comparisons. Each subtable reports $\chi^2$ and p-values; * indicates significance.}
    \label{fig:mcnemar-all}
\end{table}

\begin{table}[t]
\centering
\footnotesize
\begin{tabular}{p{1.2cm}cccc}
\toprule
\textbf{Prompt strategy} & \textbf{$H$} & \textbf{$p$-value} & \textbf{$\varepsilon^2$} & \textbf{Sig.} \\
\midrule
\multicolumn{5}{c}{\textbf{CCR\,+\,ECI\,+\,RC \textit{member}} ($k=2, n=512$)} \\
\midrule
0\_4   & $47.9251$ & $4.428\times 10^{-12}$ & $0.0920$ & * \\
0\_5   & $73.6743$ & $9.213\times 10^{-18}$ & $0.1425$ & * \\
0\_6   & $84.7026$ & $3.468\times 10^{-20}$ & $0.1641$ & * \\
F\_4   & $50.4575$ & $1.218\times 10^{-12}$ & $0.0970$ & * \\
F\_5   & $9.5520$  & $0.001997$             & $0.0168$ & * \\
F\_6   & $46.1151$ & $1.115\times 10^{-11}$ & $0.0885$ & * \\
CoT\_4 & $12.0326$ & $0.0005228$            & $0.0216$ & * \\
CoT\_5 & $13.2954$ & $0.0002661$            & $0.0241$ & * \\
CoT\_6 & $52.4552$ & $4.402\times 10^{-13}$ & $0.1009$ & * \\
\midrule
\multicolumn{5}{c}{\textbf{CCR\,+\,ECI\,+\,RC \textit{position}} ($k=4, n=512$)} \\
\midrule
0\_4   & $15.7083$ & $0.001301$             & $0.0250$ & * \\
0\_5   & $4.6300$  & $0.201$                & $0.0032$ &  \\
0\_6   & $6.2238$  & $0.1012$               & $0.0063$ &  \\
F\_4   & $6.7837$  & $0.07912$              & $0.0074$ &  \\
F\_5   & $47.0067$ & $3.464\times 10^{-10}$ & $0.0866$ & * \\
F\_6   & $29.7173$ & $1.583\times 10^{-6}$  & $0.0526$ & * \\
CoT\_4 & $7.9218$  & $0.04765$              & $0.0097$ & * \\
CoT\_5 & $7.5165$  & $0.05714$              & $0.0089$ &  \\
CoT\_6 & $28.1288$ & $3.413\times 10^{-6}$  & $0.0495$ & * \\
\bottomrule
\end{tabular}
\caption{Kruskal--Wallis test results for total complexity $C(s)$ across CCR+ECI+RC member and position label conditions. Reported are the test statistic ($H$), $p$-value, effect size ($\varepsilon^2$), and significance indicator (* for $p<0.05$).}
\label{tab:kw-total-complexity}
\end{table}

\section{On the Use of AI Assistants in Coding and Writing}

In this research, artificial intelligence assistants were used to assist in coding (Copilot) and writing (ChatGPT). After using these tools/services, the authors reviewed and edited the content as needed and take full responsibility for the content of the publication.

\begin{table*}[ht!]
    \centering
    \footnotesize
    \begin{tabular}{p{1.5cm}p{11cm}p{2.5cm}}
        \toprule
        \textbf{Variant} & \textbf{Prompt} & \textbf{Expected output} \\
        \midrule
        CorrI-0-1 & ``[MASK] in \{positive, negative\}. There is a [MASK] correlation between $e_i$ and $e_j$.'' & $\{\text{positive},\text{negative}\}$ \\
        CorrI-0-2 & ``[MASK] in \{same, opposite\}. $e_i$ impact(s) $e_j$. When we would intervene in $e_i$, $e_i$ and $e_j$ change in the [MASK] direction.'' & $\{\text{same},\text{opposite}\}$ \\
        CorrI-0-3 & ``[MASK] in \{increase, decrease\}. If $e_i$ (the cause) were to increase, $e_j$ (the effect) would [MASK].'' & $\{\text{increase},\text{decrease}\}$ \\
        CorrI-F-1 & ``There is a positive correlation between access to finance and climate resilient development. There is a negative correlation between persistent barriers and political feasibility of deploying AFOLU mitigation options. There is a [MASK] correlation between $e_i$ and $e_j$.'' & $\{\text{positive},\text{negative}\}$ \\
        CorrI-F-2 & ``Access to finance impacts climate resilient development. When we would intervene in access to finance, access to finance and climate resilient development change in the same direction. Persistent barriers impact political feasibility of deploying AFOLU mitigation options. When we would intervene in persistent barriers, persistent barriers and political feasibility of deploying AFOLU mitigation options change in the opposite direction. $e_i$ impact(s) $e_j$. When we would intervene in $e_i$, $e_i$ and $e_j$ change in the [MASK] direction.'' & $\{\text{same},\text{opposite}\}$ \\
        CorrI-F-3 & ``If access to finance (the cause) were to increase, climate resilient development (the effect) would increase. If persistent barriers (the cause) were to increase, political feasibility of deploying AFOLU mitigation options (the effect) would decrease. If $e_i$ (the cause) were to increase, $e_j$ (the effect) would [MASK].'' & $\{\text{increase},\text{decrease}\}$ \\
        CorrI-CoT-1 & ``Let's think step by step. [MASK] in \{positive, negative\}. There is a [MASK] correlation between $e_i$ and $e_j$.'' & $\{\text{positive},\text{negative}\}$ \\
        CorrI-CoT-2 & ``Let's think step by step. [MASK] in \{same, opposite\}. $e_i$ impact(s) $e_j$. When we would intervene in $e_i$, $e_i$ and $e_j$ change in the [MASK] direction.'' & $\{\text{same},\text{opposite}\}$ \\
        CorrI-CoT-3 & ``Let's think step by step. [MASK] in \{increase, decrease\}. If $e_i$ (the cause) were to increase, $e_j$ (the effect) would [MASK].'' & $\{\text{increase},\text{decrease}\}$ \\
        \bottomrule
    \end{tabular}
    \caption{Prompts for correlation inference without explicit statement context (CorrI).}
    \label{tab:prompts_corrI}
\end{table*}

\begin{table*}[ht!]
    \centering
    \footnotesize
    \begin{tabular}{p{2.5cm}p{10cm}p{2.5cm}}
        \toprule
        \textbf{Variant} & \textbf{Prompt} & \textbf{Expected output} \\
        \midrule
        CorrI+RC-0-1 & ``[MASK] in \{positive, negative\}. Statement: $s$. The statement reports a [MASK] correlation between $e_i$ and $e_j$.'' & $\{\text{positive},\text{negative}\}$ \\
        CorrI+RC-0-2 & ``[MASK] in \{same, opposite\}. Statement: $s$. $e_i$ impact(s) $e_j$. Based on the statement, when we would intervene in $e_i$, $e_i$ and $e_j$ change in the [MASK] direction.'' & $\{\text{same},\text{opposite}\}$ \\
        CorrI+RC-0-3 & ``[MASK] in \{increase, decrease\}. Given the statement: $s$. If $e_i$ (the cause) were to increase, $e_j$ (the effect) would [MASK].'' & $\{\text{increase},\text{decrease}\}$ \\
        CorrI+RC-F-1 & ``Statement: ``Climate resilient development is enabled by increased international cooperation including mobilising and enhancing access to finance, particularly for developing countries, vulnerable regions, sectors and groups and aligning finance flows for climate action to be consistent with ambition levels and funding needs.''. The statement reports a positive correlation between access to finance and climate resilient development. Statement: ``Persistent and region-specific barriers also continue to hamper the economic and political feasibility of deploying AFOLU mitigation options.''. The statement reports a negative correlation between persistent barriers and political feasibility of deploying AFOLU mitigation options. Statement: $s$. The statement reports a [MASK] correlation between $e_i$ and $e_j$.'' & $\{\text{positive},\text{negative}\}$ \\
        CorrI+RC-F-2 & ``Statement: ``Climate resilient development is enabled by increased international cooperation including mobilising and enhancing access to finance, particularly for developing countries, vulnerable regions, sectors and groups and aligning finance flows for climate action to be consistent with ambition levels and funding needs.''. Access to finance impacts climate resilient development. Based on the statement, when we would intervene in access to finance, access to finance and climate resilient development change in the same direction. Statement: ``Persistent and region-specific barriers also continue to hamper the economic and political feasibility of deploying AFOLU mitigation options.''. Persistent barriers impact political feasibility of deploying AFOLU mitigation options. Based on the statement, when we would intervene in persistent barriers, persistent barriers and political feasibility of deploying AFOLU mitigation options change in the opposite direction. Statement: $s$. $e_i$ impact(s) $e_j$. Based on the statement, when we would intervene in $e_i$, $e_i$ and $e_j$ change in the [MASK] direction.'' & $\{\text{same},\text{opposite}\}$ \\
        CorrI+RC-F-3 & ``Given the statement: ``Climate resilient development is enabled by increased international cooperation including mobilising and enhancing access to finance, particularly for developing countries, vulnerable regions, sectors and groups and aligning finance flows for climate action to be consistent with ambition levels and funding needs.''. If access to finance (the cause) were to increase, climate resilient development (the effect) would increase. Given the statement: ``Persistent and region-specific barriers also continue to hamper the economic and political feasibility of deploying AFOLU mitigation options.''. If persistent barriers (the cause) were to increase, political feasibility of deploying AFOLU mitigation options (the effect) would decrease. Given the statement: $s$. If $e_i$ (the cause) were to increase, $e_j$ (the effect) would [MASK].'' & $\{\text{increase},\text{decrease}\}$ \\
        CorrI+RC-CoT-1 & ``Let's think step by step. [MASK] in \{positive, negative\}. Statement: $s$. The statement reports a [MASK] correlation between $e_i$ and $e_j$.'' & $\{\text{positive},\text{negative}\}$ \\
        CorrI+RC-CoT-2 & ``Let's think step by step. [MASK] in \{same, opposite\}. Statement: $s$. $e_i$ impact(s) $e_j$. Based on the statement, when we would intervene in $e_i$, $e_i$ and $e_j$ change in the [MASK] direction.'' & $\{\text{same},\text{opposite}\}$ \\
        CorrI+RC-CoT-3 & ``Let's think step by step. Statement: $s$. [MASK] in \{increase, decrease\}. If $e_i$ (the cause) were to increase, $e_j$ (the effect) would [MASK].'' & $\{\text{increase},\text{decrease}\}$ \\
        \bottomrule
    \end{tabular}
    \caption{Prompts for correlation inference with explicit statement context.}
    \label{tab:prompts_corrI_RC}
\end{table*}

\begin{table*}[ht!]
    \centering
    \footnotesize
    \begin{tabular}{p{2cm}p{11cm}p{2cm}}
        \toprule
        \textbf{Variant} & \textbf{Prompt} & \textbf{Expected output} \\
        \midrule
        CCR\_member\_ A\_4 & ``You will be given a causal graph. The causal relationships in this causal graph are- $G_A$. Now answer using this causal graph only, determine whether $e_i$ is part of a causal chain. A causal chain is a directed path of at least three nodes in a causal graph. Think step by step. Give reasoning and then give an answer within \textless Answer\textgreater \{Yes, No\} \textless/Answer\textgreater.'' & \{Yes, No\} \\
        CCR\_member\_ A\_5 & ``The causal relationships in a causal graph are- $G_A$. Based on this graph, determine whether $e_i$ belongs to a causal chain. A causal chain is a directed path of at least three nodes in a causal graph. Answer with \{Yes, No\} only.'' & \{Yes, No\} \\
        CCR\_member\_ A\_6 & ``The given causal graph includes the following causal relations: $G_A$. Study this graph carefully, and decide whether the graph contains a causal chain structure that includes $e_i$. Answer with \{Yes, No\} only.'' & \{Yes, No\} \\
        CCR\_member\_ SN\_4 & ``You will be given a causal graph. The causal relationships in this causal graph are- $G_{SN}$. Now answer using this causal graph only, determine whether $e_i$ is part of a causal chain. A causal chain is a directed path of at least three nodes in a causal graph. Think step by step. Give reasoning and then give an answer within \textless Answer\textgreater \{Yes, No\} \textless/Answer\textgreater.'' & \{Yes, No\} \\
        CCR\_member\_ SN\_5 & ``The causal relationships in a causal graph are- $G_{SN}$. Based on this graph, determine whether $e_i$ belongs to a causal chain. A causal chain is a directed path of at least three nodes in a causal graph. Answer with \{Yes, No\} only.'' & \{Yes, No\} \\
        CCR\_member\_ SN\_6 & ``The given causal graph includes the following causal relations: $G_{SN}$. Study this graph carefully, and decide whether the graph contains a causal chain structure that includes $e_i$. Answer with \{Yes, No\} only.'' & \{Yes, No\} \\
        CCR\_member\_ ML\_4 & ``You will be given a causal graph. The causal relationships in this causal graph are- $G_{ML}$. Now answer using this causal graph only, determine whether $e_i$ is part of a causal chain. A causal chain is a directed path of at least three nodes in a causal graph. Think step by step. Give reasoning and then give an answer within \textless Answer\textgreater \{Yes, No\} \textless/Answer\textgreater.'' & \{Yes, No\} \\
        CCR\_member\_ ML\_5 & ``The causal relationships in a causal graph are- $G_{ML}$. Based on this graph, determine whether $e_i$ belongs to a causal chain. A causal chain is a directed path of at least three nodes in a causal graph. Answer with \{Yes, No\} only.'' & \{Yes, No\} \\
        CCR\_member\_ ML\_6 & ``The given causal graph includes the following causal relations: $G_{ML}$. Study this graph carefully, and decide whether the graph contains a causal chain structure that includes $e_i$. Answer with \{Yes, No\} only.'' & \{Yes, No\} \\
        \bottomrule
    \end{tabular}
    \caption{Prompts for causal chain reasoning (CCR) benchmarking (\textbf{\textit{membership}}), where $e_i$ is an entity and $G_A$, $G_{SN}$, $G_{ML}$ represent causal graphs.}
    \label{tab:prompts_ccr}
\end{table*}

\begin{table*}[ht!]
    \centering
    \footnotesize
    \begin{tabular}{p{2cm}p{10.5cm}p{2.5cm}}
        \toprule
        \textbf{Variant} & \textbf{Prompt} & \textbf{Expected output} \\
        \midrule
        CCR\_position\_ A\_4 & ``You will be given a causal graph. The causal relationships in this causal graph are- $G_A$. Now answer using this causal graph only, determine whether $e_i$ is part of a causal chain and, if yes, which position it holds in that chain. A causal chain is a directed path of at least three nodes in a causal graph. Think step by step. Give reasoning and then give an answer within \textless Answer\textgreater \{start, middle, end, none\} \textless/Answer\textgreater.'' & \{start, middle, end, none\} \\
        CCR\_position\_ A\_5 & ``The causal relationships in a causal graph are- $G_A$. Based on this graph, determine whether $e_i$ belongs to a causal chain and, if yes, which position in the chain that event can be found (start, middle, or end). A causal chain is a directed path of at least three nodes in a causal graph. Answer with \{start, middle, end, none\} only.'' & \{start, middle, end, none\} \\
        CCR\_position\_ A\_6 & ``The given causal graph includes the following causal relations: $G_A$. Study this graph carefully, and decide whether the graph contains a causal chain structure that includes $e_i$ and which position $e_i$ holds in that chain. Answer with \{start, middle, end, none\} only.'' & \{start, middle, end, none\} \\
        CCR\_position\_ SN\_4 & ``You will be given a causal graph. The causal relationships in this causal graph are- $G_{SN}$. Now answer using this causal graph only, determine whether $e_i$ is part of a causal chain and, if yes, which position it holds in that chain. A causal chain is a directed path of at least three nodes in a causal graph. Think step by step. Give reasoning and then give an answer within \textless Answer\textgreater \{start, middle, end, none\} \textless/Answer\textgreater.'' & \{start, middle, end, none\} \\
        CCR\_position\_ SN\_5 & ``The causal relationships in a causal graph are- $G_{SN}$. Based on this graph, determine whether $e_i$ belongs to a causal chain and, if yes, which position in the chain that event can be found (start, middle, or end). A causal chain is a directed path of at least three nodes in a causal graph. Answer with \{start, middle, end, none\} only.'' & \{start, middle, end, none\} \\
        CCR\_position\_ SN\_6 & ``The given causal graph includes the following causal relations: $G_{SN}$. Study this graph carefully, and decide whether the graph contains a causal chain structure that includes $e_i$ and which position $e_i$ holds in that chain. Answer with \{start, middle, end, none\} only.'' & \{start, middle, end, none\} \\
        CCR\_position\_ ML\_4 & ``You will be given a causal graph. The causal relationships in this causal graph are- $G_{ML}$. Now answer using this causal graph only, determine whether $e_i$ is part of a causal chain and, if yes, which position it holds in that chain. A causal chain is a directed path of at least three nodes in a causal graph. Think step by step. Give reasoning and then give an answer within \textless Answer\textgreater \{start, middle, end, none\} \textless/Answer\textgreater.'' & \{start, middle, end, none\} \\
        CCR\_position\_ ML\_5 & ``The causal relationships in a causal graph are- $G_{ML}$. Based on this graph, determine whether $e_i$ belongs to a causal chain and, if yes, which position in the chain that event can be found (start, middle, or end). A causal chain is a directed path of at least three nodes in a causal graph. Answer with \{start, middle, end, none\} only.'' & \{start, middle, end, none\} \\
        CCR\_position\_ ML\_6 & ``The given causal graph includes the following causal relations: $G_{ML}$. Study this graph carefully, and decide whether the graph contains a causal chain structure that includes $e_i$ and which position $e_i$ holds in that chain. Answer with \{start, middle, end, none\} only.'' & \{start, middle, end, none\} \\
        \bottomrule
    \end{tabular}
    \caption{Prompts for causal chain reasoning (CCR) benchmarking (\textbf{\textit{position}}), where $e_i$ is an entity and $G_A$, $G_{SN}$, $G_{ML}$ represent causal graphs.}
    \label{tab:prompts_ccr_position}
\end{table*}

\begin{table*}[ht!]
    \centering
    \footnotesize
    \begin{tabular}{p{2cm}p{11.8cm}p{1.3cm}}
        \toprule
        \textbf{Variant} & \textbf{Prompt} & \textbf{Expected output} \\
        \midrule
        CCR\_ECI\_ member\_0\_4 & ``Statement: ``$s$''. All causal events in the statement: $event\_list$. Based on the statement and the provided list of causal events, determine whether $e_i$ is reported as part of a causal chain. A causal chain is a directed path of at least three nodes in a causal graph. Give an answer within \textless Answer\textgreater \{Yes, No\} \textless/Answer\textgreater.'' & \{Yes, No\} \\
        CCR\_ECI\_ member\_0\_5 & ``Given a statement: ``$s$'' and a list of all causal events: $event\_list$. Determine whether $e_i$ belongs to a causal chain reported, either explicitly or implicitly, in the statement. A causal chain is a directed path of at least three nodes in a causal graph. Answer with \{Yes, No\} only.'' & \{Yes, No\} \\
        CCR\_ECI\_ member\_0\_6 & ``The following list of causal events $event\_list$ can be found in statement ``$s$''. Does the statement report a causal chain structure that includes $e_i$? A causal chain is a directed path of at least three nodes in a causal graph. Answer with \{Yes, No\} only.'' & \{Yes, No\} \\
        CCR\_ECI\_ member\_F\_4 & ``A causal chain is a directed path of at least three nodes in a causal graph. Statement: ``Persistent and region-specific barriers also continue to hamper the economic and political feasibility of deploying AFOLU mitigation options.''. All causal events in the statement: [persistent barriers, region-specific barriers, economic feasibility of deploying AFOLU mitigation options, political feasibility of deploying AFOLU mitigation options]. Based on the statement and the provided list of causal events, is region-specific barriers reported as part of a causal chain? No. Statement: ``Adaptation options can become maladaptive due to their environmental impacts that constrain ecosystem services and decrease biodiversity and ecosystem resilience to climate change or by causing adverse outcomes for different groups, exacerbating inequity.''. All causal events in the statement: [adaptation options, environmental impacts, ecosystem services, biodiversity, ecosystem resilience to climate change, adverse outcomes for different groups, inequity, maladaptive adaptation options]. Based on the statement and the provided list of causal events, is inequity reported as part of a causal chain? Yes. Statement: ``$s$''. All causal events in the statement: $event\_list$. Based on the statement and the provided list of causal events, is $e_i$ reported as part of a causal chain?'' & \{Yes, No\} \\
        CCR\_ECI\_ member\_F\_5 & ``A causal chain is a directed path of at least three nodes in a causal graph. Given a statement: ``Persistent and region-specific barriers also continue to hamper the economic and political feasibility of deploying AFOLU mitigation options.'' and a list of all causal events: [persistent barriers, region-specific barriers, economic feasibility of deploying AFOLU mitigation options, political feasibility of deploying AFOLU mitigation options]. Determine whether region-specific barriers belongs to a causal chain reported, either explicitly or implicitly, in the statement. Answer: No. Given a statement: ``Adaptation options can become maladaptive due to their environmental impacts that constrain ecosystem services and decrease biodiversity and ecosystem resilience to climate change or by causing adverse outcomes for different groups, exacerbating inequity.'' and a list of all causal events: [adaptation options, environmental impacts, ecosystem services, biodiversity, ecosystem resilience to climate change, adverse outcomes for different groups, inequity, maladaptive adaptation options]. Determine whether inequity belongs to a causal chain reported, either explicitly or implicitly, in the statement. Answer: Yes. Given a statement: ``$s$'' and a list of all causal events: $event\_list$. Determine whether $e_i$ belongs to a causal chain reported, either explicitly or implicitly, in the statement. Answer:'' & \{Yes, No\} \\
        CCR\_ECI\_ member\_F\_6 & ``A causal chain is a directed path of at least three nodes in a causal graph. The following list of causal events [persistent barriers, region-specific barriers, economic feasibility of deploying AFOLU mitigation options, political feasibility of deploying AFOLU mitigation options] can be found in statement ``Persistent and region-specific barriers also continue to hamper the economic and political feasibility of deploying AFOLU mitigation options.''. Does the statement report a causal chain structure that includes region-specific barriers? No. The following list of causal events [adaptation options, environmental impacts, ecosystem services, biodiversity, ecosystem resilience to climate change, adverse outcomes for different groups, inequity, maladaptive adaptation options] can be found in statement ``Adaptation options can become maladaptive due to their environmental impacts that constrain ecosystem services and decrease biodiversity and ecosystem resilience to climate change or by causing adverse outcomes for different groups, exacerbating inequity.''. Does the statement report a causal chain structure that includes inequity? Yes. The following list of causal events $event\_list$ can be found in statement ``$s$''. Does the statement report a causal chain structure that includes $e_i$?'' & \{Yes, No\} \\
        CCR\_ECI\_ member\_CoT\_4 & ``Statement: ``$s$''. All causal events in the statement: $event\_list$. Based on the statement and the provided list of causal events, determine whether $e_i$ is reported as part of a causal chain. A causal chain is a directed path of at least three nodes in a causal graph. Think step by step. Give reasoning and then give an answer within \textless Answer\textgreater \{Yes, No\} \textless/Answer\textgreater.'' & \{Yes, No\} \\
        CCR\_ECI\_ member\_CoT\_5 & ``Given a statement: ``$s$'' and a list of all causal events: $event\_list$. Determine whether $e_i$ belongs to a causal chain reported, either explicitly or implicitly, in the statement. A causal chain is a directed path of at least three nodes in a causal graph. Think step by step. Finally, answer with \{Yes, No\}.'' & \{Yes, No\} \\
        CCR\_ECI\_ member\_CoT\_6 & ``The following list of causal events $event\_list$ can be found in statement ``$s$''. Does the statement report a causal chain structure that includes $e_i$? A causal chain is a directed path of at least three nodes in a causal graph. Let's think step by step and answer with \{Yes, No\}.'' & \{Yes, No\} \\
        \bottomrule
    \end{tabular}
    \caption{Prompts for causal reasoning benchmarking (CCR+ECI+RC) (\textbf{\textit{membership}}), where $s$ is a statement and $event\_list$ is the list of causal events extracted from it.}
    \label{tab:prompts_ccr_eci}
\end{table*}

\begin{table*}[ht!]
    \centering
    \footnotesize
    \begin{tabular}{p{2cm}p{11.8cm}p{1.3cm}}
        \toprule
        \textbf{Variant} & \textbf{Prompt} & \textbf{Expected output} \\
        \midrule
        CCR\_ECI\_ position\_0\_4 & ``Statement: ``$s$''. All causal events in the statement: $event\_list$. Based on the statement and the provided list of causal events, determine whether $e_i$ is reported as part of a causal chain and, if yes, which position it holds in that chain. A causal chain is a directed path of at least three nodes in a causal graph. Give an answer within \textless Answer\textgreater \{start, middle, end, none\} \textless/Answer\textgreater.'' & \{start, middle, end, none\} \\
        CCR\_ECI\_ position\_0\_5 & ``Given a statement: ``$s$'' and a list of all causal events: $event\_list$. Determine whether $e_i$ belongs to a causal chain reported, either explicitly or implicitly, in the statement and which position it holds in that chain. A causal chain is a directed path of at least three nodes in a causal graph. Answer with \{start, middle, end, none\} only.'' & \{start, middle, end, none\} \\
        CCR\_ECI\_ position\_0\_6 & ``The following list of causal events $event\_list$ can be found in statement ``$s$''. Does the statement report a causal chain structure that includes $e_i$ and at which position in the chain can the event be found? A causal chain is a directed path of at least three nodes in a causal graph. Answer with \{start, middle, end, none\} only.'' & \{start, middle, end, none\} \\
        CCR\_ECI\_ position\_F\_4 & ``A causal chain is a directed path of at least three nodes in a causal graph. Statement: ``Persistent and region-specific barriers also continue to hamper the economic and political feasibility of deploying AFOLU mitigation options.''. All causal events in the statement: [persistent barriers, region-specific barriers, economic feasibility of deploying AFOLU mitigation options, political feasibility of deploying AFOLU mitigation options]. Based on the statement and the provided list of causal events, is region-specific barriers reported as part of a causal chain, and, if yes, which position it holds in that chain? none. Statement: ``Adaptation options can become maladaptive due to their environmental impacts that constrain ecosystem services and decrease biodiversity and ecosystem resilience to climate change or by causing adverse outcomes for different groups, exacerbating inequity.''. All causal events in the statement: [adaptation options, environmental impacts, ecosystem services, biodiversity, ecosystem resilience to climate change, adverse outcomes for different groups, inequity, maladaptive adaptation options]. Based on the statement and the provided list of causal events, is inequity reported as part of a causal chain, and, if yes, which position it holds in that chain? middle. Statement: ``$s$''. All causal events in the statement: $event\_list$. Based on the statement and the provided list of causal events, is $e_i$ reported as part of a causal chain, and, if yes, which position it holds in that chain? Answer with \{start, middle, end, none\} only.'' & \{start, middle, end, none\} \\
        CCR\_ECI\_ position\_F\_5 & ``A causal chain is a directed path of at least three nodes in a causal graph. Given a statement: ``Persistent and region-specific barriers also continue to hamper the economic and political feasibility of deploying AFOLU mitigation options.'' and a list of all causal events: [persistent barriers, region-specific barriers, economic feasibility of deploying AFOLU mitigation options, political feasibility of deploying AFOLU mitigation options]. Determine whether region-specific barriers belongs to a causal chain reported, either explicitly or implicitly, in the statement and answer with the position the event holds in that chain. Answer: none. Given a statement: ``Adaptation options can become maladaptive due to their environmental impacts that constrain ecosystem services and decrease biodiversity and ecosystem resilience to climate change or by causing adverse outcomes for different groups, exacerbating inequity.'' and a list of all causal events: [adaptation options, environmental impacts, ecosystem services, biodiversity, ecosystem resilience to climate change, adverse outcomes for different groups, inequity, maladaptive adaptation options]. Determine whether inequity belongs to a causal chain reported, either explicitly or implicitly, in the statement and answer with the position the event holds in that chain. Answer: middle. Given a statement: ``$s$'' and a list of all causal events: $event\_list$. Determine whether $e_i$ belongs to a causal chain reported, either explicitly or implicitly, in the statement and answer with the position the event holds in that chain. \{start, middle, end, none\} Answer:'' & \{start, middle, end, none\} \\
        CCR\_ECI\_ position\_F\_6 & ``A causal chain is a directed path of at least three nodes in a causal graph. Answer with \{start, middle, end, none\} only. The following list of causal events [persistent barriers, region-specific barriers, economic feasibility of deploying AFOLU mitigation options, political feasibility of deploying AFOLU mitigation options] can be found in statement ``Persistent and region-specific barriers also continue to hamper the economic and political feasibility of deploying AFOLU mitigation options.''. Does the statement report a causal chain structure that includes region-specific barriers and what position does region-specific barriers hold? none. The following list of causal events [adaptation options, environmental impacts, ecosystem services, biodiversity, ecosystem resilience to climate change, adverse outcomes for different groups, inequity, maladaptive adaptation options] can be found in statement ``Adaptation options can become maladaptive due to their environmental impacts that constrain ecosystem services and decrease biodiversity and ecosystem resilience to climate change or by causing adverse outcomes for different groups, exacerbating inequity.''. Does the statement report a causal chain structure that includes inequity and what position does inequity hold? middle. The following list of causal events $event\_list$ can be found in statement ``$s$''. Does the statement report a causal chain structure that includes $e_i$ and what position does $e_i$ hold?'' & \{start, middle, end, none\} \\
        \bottomrule
    \end{tabular}
    \caption{Prompts for causal reasoning benchmarking (CCR+ECI+RC) (\textbf{\textit{position}}) – examples and zero-shot variants.}
    \label{tab:prompts_ccr_eci_position_part1}
\end{table*}

\begin{table*}[ht!]
    \centering
    \footnotesize
    \begin{tabular}{p{2cm}p{11.8cm}p{1.3cm}}
        \toprule
        \textbf{Variant} & \textbf{Prompt} & \textbf{Expected output} \\
        \midrule
        CCR\_ECI\_ position\_CoT\_4 & ``Statement: ``$s$''. All causal events in the statement: $event\_list$. Based on the statement and the provided list of causal events, determine whether $e_i$ is reported as part of a causal chain and, if yes, which position it holds in that chain. A causal chain is a directed path of at least three nodes in a causal graph. Think step by step. Give reasoning and then give an answer within \textless Answer\textgreater \{start, middle, end, none\} \textless/Answer\textgreater.'' & \{start, middle, end, none\} \\
        CCR\_ECI\_ position\_CoT\_5 & ``Given a statement: ``$s$'' and a list of all causal events: $event\_list$. Determine whether $e_i$ belongs to a causal chain reported, either explicitly or implicitly, in the statement and which position it holds in that chain. A causal chain is a directed path of at least three nodes in a causal graph. Think step by step. Finally, answer with \{start, middle, end, none\}.'' & \{start, middle, end, none\} \\
        CCR\_ECI\_ position\_CoT\_6 & ``The following list of causal events $event\_list$ can be found in statement ``$s$''. Does the statement report a causal chain structure that includes $e_i$ and at which position in the chain can the event be found? A causal chain is a directed path of at least three nodes in a causal graph. Let's think step by step and answer with \{start, middle, end, none\}.'' & \{start, middle, end, none\} \\
        \bottomrule
    \end{tabular}
    \caption{Prompts for causal reasoning benchmarking (CCR+ECI+RC) (\textbf{\textit{position}}) – CoT variants.}
    \label{tab:prompts_ccr_eci_position_part2}
\end{table*}

\end{document}